# Communication-Avoiding Optimization Methods for Distributed Massive-Scale Sparse Inverse Covariance Estimation

Penporn Koanantakool[1,3,*], Alnur Ali[2], Ariful Azad[3], Aydın Buluç[1,3],
Dmitriy Morozov[3,4], Leonid Oliker[3], Katherine Yelick[1,3], Sang-Yun Oh[3,5]

[1]Department of Electrical Engineering and Computer Sciences, UC Berkeley
[2]Machine Learning Department, Carnegie Mellon University
[3]Computational Research Division, Lawrence Berkeley National Laboratory
[4]Berkeley Institute for Data Science, UC Berkeley
[5]Department of Statistics and Applied Probability, UC Santa Barbara

## Abstract

Across a variety of scientific disciplines, sparse inverse covariance estimation is a popular tool for capturing the underlying dependency relationships in multivariate data. Unfortunately, most estimators are not scalable enough to handle the sizes of modern high-dimensional data sets (often on the order of terabytes), and assume Gaussian samples. To address these deficiencies, we introduce HP-CONCORD, a highly scalable optimization method for estimating a sparse inverse covariance matrix based on a regularized *pseudolikelihood* framework, without assuming Gaussianity. Our parallel proximal gradient method uses a novel *communication-avoiding* linear algebra algorithm and runs across a multi-node cluster with up to 1k nodes (24k cores), achieving parallel scalability on problems with up to ≈819 billion parameters (1.28 million dimensions); even on a single node, HP-CONCORD demonstrates scalability, outperforming a state-of-the-art method. We also use HP-CONCORD to estimate the underlying dependency structure of the brain from fMRI data, and use the result to identify functional regions automatically. The results show good agreement with a clustering from the neuroscience literature.

## 1 Introduction

Suppose we observe $n$ samples of a $p$-dimensional random vector $(X_1, \ldots, X_p)$, drawn independently and identically from an unknown distribution, which we assume without loss of generality has mean zero and covariance matrix $\Sigma^0 \in \mathbf{S}^p_{++}$, where $\mathbf{S}^p_{++}$ denotes the space of $(p \times p)$-dimensional positive definite matrices. In this paper, we are interested in obtaining a sparse estimate of the underlying inverse covariance matrix $\Omega^0 = (\Sigma^0)^{-1}$ associated with the $p$ random variables, even in a high-dimensional setup, where it may be the case that $p \gg n$. This problem, known as sparse inverse covariance estimation, is an important one in statistics, and plays a role in a wide variety of real-world applications, including finance (*e.g.*, [34, 32, 53, 27, 4]), biology (*e.g.*, [22, 27, 38]), and sustainability (*e.g.*, [54, 4, 3]).

---

[*]Now at Google Brain.

There are at least two reasons for the popularity of sparse inverse covariance estimation. First, it is well-known that the sparsity pattern of the underlying inverse covariance matrix $\Omega^0$ gives rise to a *partial correlation graph* associated with the random variables $X_1, \ldots, X_p$; thus, an estimate of $\Omega^0$ can reveal the statistical relationships between the variables. To be more concrete: we construct an undirected graph, where the vertices correspond to the variables $X_1, \ldots, X_p$, and we put an edge between two vertices if and only if their (estimated) partial correlation coefficient is nonzero. Under the assumption that the underlying data-generating distribution is multivariate normal, the partial correlation graph is precisely the conditional independence graph associated with the variables [31, 30, 8]. It also turns out that many downstream applications readily use an estimate of the inverse covariance matrix, not just its sparsity pattern.

The literature on sparse inverse covariance estimation is vast, so we cannot possibly give a complete coverage of it here; some influential papers include [36, 56, 22, 43, 10]. On the computational side, a lot of the recent work has looked at developing scalable algorithms, specifically for a shared memory environment. However, in a "massive-scale" setting, where $p, n$ are so large that using a single machine/node is infeasible, it may be more suitable to consider a distributed memory approach; *e.g.*, functional magnetic resonance imaging (fMRI) data sets easily run into the hundreds of gigabytes and require fitting billions of parameters. In these cases, the literature is somewhat lacking, with a few exceptions that we return to later [51, 26].

With this motivation in mind, we propose a new, highly scalable parallel proximal gradient method, for obtaining a sparse estimate of the inverse covariance matrix, in shared and/or distributed memory settings. The method, called HP-CONCORD ("HP" stands for "high-performance"), builds on the recently introduced CONCORD [27, 38] and PseudoNet [4] estimators, and explicitly minimizes the communication costs between nodes by leveraging ideas from the literature on *communication-avoiding algorithms* [16]. Highlighting some of our findings: on a single node, HP-CONCORD is about an order of magnitude faster at fitting ≈800 million parameters than BigQUIC, a well-known method for scalable sparse inverse covariance estimation [25], and also demonstrates good scalability on a cluster with 1,024 nodes, where it is able to fit ≈819 billion parameters in ≈17 minutes.

Here is an outline for the rest of the paper. In the next section, we give background on the CONCORD and PseudoNet estimators, as well as communication-avoiding algorithms, required to understand our method; in Section 3, we describe our method, HP-CONCORD. In Section 4, we evaluate HP-CONCORD on high-dimensional synthetic data, comparing it to BigQUIC. In Section 5, we present an in-depth empirical study, where we use HP-CONCORD to estimate the underlying partial correlation structure of the brain from high-dimensional fMRI data, requiring fitting ≈4 billion parameters. We wrap-up in Section 6.

## 2 Background

**CONCORD.** Recent work [27] proposed the CONCORD estimator; CONCORD is a *pseudo-likelihood*-based [12] estimator of the inverse covariance matrix, meaning (roughly) that it obtains an estimate by solving a sequence of lasso-like problems, rather than explicitly minimizing an $\ell_1$-penalized Gaussian likelihood, making CONCORD suitable for situations where the underlying distribution is suspected to be non-Gaussian. Along these lines, CONCORD outperformed a number of strong competitors, including the graphical lasso [22], on several real-world data sets [27, 38, 4]. CONCORD also enjoys favorable estimation error and support recovery guarantees [27, 4].

**PseudoNet.** Follow-up work [4] proposed the PseudoNet estimator, which generalizes CONCORD, and attains much better statistical and empirical performance. The PseudoNet estimate is defined as the solution to the convex optimization problem,

$$\underset{\Omega \in \mathbf{R}^{p \times p}}{\text{minimize}} \quad -\log \det(\Omega_D^2) + \mathbf{tr}(\Omega S \Omega) + \lambda_1 \|\Omega_X\|_1 + \frac{\lambda_2}{2} \|\Omega\|_F^2, \tag{1}$$

where $\Omega_D, \Omega_X \in \mathbf{R}^{p \times p}$ denote the matrices containing just the diagonal and off-diagonal entries of $\Omega$, respectively; $S = \frac{1}{n} X^T X \in \mathbf{S}_+^p$ is the sample covariance matrix; $X \in \mathbf{R}^{n \times p}$ is the observation matrix; $\lambda_1, \lambda_2 \geq 0$ are tuning parameters; and $\|\cdot\|_1, \|\cdot\|_F$ denote the elementwise $\ell_1$- and Frobenius norms, respectively. As far as the criterions are concerned, the only difference between PseudoNet and CONCORD is the presence of the squared Frobenius norm penalty, *i.e.*, setting $\lambda_2 = 0$ recovers the CONCORD criterion, analogous to the relationship between the elastic net [57] and the lasso. Thus, to keep things simple, we use the names CONCORD and PseudoNet interchangeably.

As the criterion (1) is the sum of smooth and nonsmooth convex functions, optimizing (1) with a proximal gradient method [39] is natural. Applying a proximal gradient method to (1), as in [4], yields Algorithm 1. A comment on notation: we use $\mathcal{S}_\alpha(Z)$ to denote the elementwise soft-thresholding operator (*i.e.*, the proximal operator of the $\ell_1$-norm) at $Z \in \mathbf{R}^{p \times p}$ with $\alpha > 0$,

$$[\mathcal{S}_\alpha(Z)]_{ij} = \begin{cases} Z_{ij} - \alpha, & Z_{ij} > \alpha \\ Z_{ij} + \alpha, & Z_{ij} < -\alpha \\ 0, & \text{otherwise} \end{cases}, \; i,j = 1, \ldots, p. \tag{2}$$

---

**Algorithm 1** Proximal gradient method for computing the CONCORD/PseudoNet estimate [4].

---

**Input:** data matrix $X \in \mathbf{R}^{n \times p}$; tuning parameters $\lambda_1, \lambda_2 \geq 0$; optimization tolerance $\epsilon > 0$
**Output:** estimate $\hat{\Omega} \in \mathbf{R}^{p \times p}$ of the underlying inverse covariance matrix $\Omega^0$
1: Set the initial point to the identity matrix: $\Omega^{(0)} \leftarrow I$
2: Compute $S \leftarrow \frac{1}{n} X^T X$ ▷ Compute the sample covariance matrix (once)
3: **for** $k = 0, 1, 2, \ldots$ ▷ $k$ denotes the iteration counter
4:     $G^{(k)} \leftarrow -(\Omega_D^{(k)})^{-1} + \frac{1}{2}(S\Omega^{(k)} + \Omega^{(k)}S) + \lambda_2 \Omega^{(k)}$ ▷ Compute the gradient, $G^{(k)}$, of the smooth part of (1)
5:     $g(\Omega^{(k)}) \leftarrow -\log \det((\Omega_D^{(k)})^2) + \mathbf{tr}(\Omega^{(k)} S \Omega^{(k)}) + \frac{\lambda_2}{2} \|\Omega^{(k)}\|_F^2$ ▷ Evaluate the smooth part of (1), used below
6:     **for** $\tau = 1, \frac{1}{2}, \frac{1}{4}, \ldots$ ▷ Choose the step size $\tau$ via backtracking line search
7:         $\Omega^{(k+1)} \leftarrow \mathcal{S}_{\tau \lambda_1}(\Omega^{(k)} - \tau G^{(k)})$ ▷ Apply the elementwise soft-thresholding operator $\mathcal{S}_{\tau \lambda_1}$; see (2)
8:         $g(\Omega^{(k+1)}) \leftarrow -\log \det((\Omega_D^{(k+1)})^2) + \mathbf{tr}(\Omega^{(k+1)} S \Omega^{(k+1)}) + \frac{\lambda_2}{2} \|\Omega^{(k+1)}\|_F^2$ ▷ Evaluate the smooth part of (1)
9:     **until** $g(\Omega^{(k+1)}) \leq g(\Omega^{(k)}) - \mathbf{tr}((\Omega^{(k)} - \Omega^{(k+1)})^T G^{(k)}) + \frac{1}{2\tau} \|\Omega^{(k)} - \Omega^{(k+1)}\|_F^2$
10: **until** a stopping criterion is satisfied, using $\epsilon$
11: **return** the estimate $\hat{\Omega} \leftarrow \Omega^{(k)}$

---

**Key bottlenecks.** Algorithm 1 has some key computational bottlenecks that are especially problematic in a high-dimensional setting. First and foremost, assuming dense matrices, computing the matrix product $\Omega^{(k)} S$, on each proximal gradient and line search iteration, costs $O(p^3)$. Computing the covariance matrix $S$ has the one-time upfront cost of $O(p^2 n)$. Finally, transposing $\Omega^{(k)} S$ swaps $O(p^2)$ entries each iteration.

Thus, despite CONCORD's many favorable statistical properties, these bottlenecks make scaling CONCORD to massive data sets challenging. For example, Algorithm 1 can reconstruct the underlying gene-gene associations in a breast cancer data set, where $p \approx 4k$, in just $\approx 10$ minutes, but it quickly becomes extremely slow or even intractable when analyzing the complete gene expression data set, where $p \approx 30$k [38]. Furthermore, the running time required to compute the CONCORD estimates across a grid of tuning parameters, as in resampling methods such as cross-validation, the bootstrap, and stability selection [37, 33], would be prohibitive. To address these computational deficiencies, we build on the recent work coming out of the literature on communication-avoiding algorithms.

**Communication-avoiding algorithms.** Communication often dominates the overall cost of a distributed algorithm [5, 19] and therefore should be minimized. In scientific computing, communication-avoiding algorithms (see, *e.g.*, [9, 46, 20]) significantly reduce communication by (i) choosing a suitable *data layout*, *i.e.*, sensibly distributing data across nodes; (ii) *replicating*, *i.e.*, duplicating data across nodes to cut down on communication at the expense of space; and (iii) computing a quantity of interest, given the layout and replication, in a communication-efficient way. Speedups ranging from 2-100× have been reported in the literature [17, 29, 28, 47, 52, 7]; in fact, communication-avoiding algorithms decrease communication asymptotically in the amount of replication. However, communication-avoiding algorithms have gone relatively unnoticed in statistics, until very recently [55, 18, 48]. Our approach is motivated by the recent "2.5D" [46] and "1.5D" [29] communication-avoiding matrix multiplication algorithms.

**Related work.** Wang et al. [50] present a distributed memory approach to sparse inverse covariance estimation, based on the alternating direction method of multipliers [14], without optimizing the amount of communication; it would be interesting to combine their method with a communication-avoiding algorithm. GINCO [26] is a distributed greedy algorithm, which is not guaranteed to converge to a globally optimal point. We compare to BigQUIC [25], a well-known Gaussian likelihood method.

## 3 HP-CONCORD

HP-CONCORD resolves the key computational bottlenecks in Algorithm 1, by distributing the data (*e.g.*, $X$), any intermediate variables (*e.g.*, the iterates $\Omega^{(k)}$), and any computations across a network of nodes/processors. Some of the computations may then be done in an embarrassingly parallel way; for the others, we turn to a communication-avoiding approach.

We begin by observing that $\Omega^{(k)}S$, critical to most of the calculations in Algorithm 1, can be computed in two ways, with different computation and communication costs. The first approach, which we call "Cov", explicitly computes $S = \frac{1}{n}X^TX$, using $S$ to then compute $\Omega^{(k)}S$. The second approach, which we call "Obs", never explicitly computes $S$, opting instead to compute $Y^{(k)} = \frac{1}{n}\Omega^{(k)}X^T$ and $Y^{(k)}X = \Omega^{(k)}S$. (The fact that Cov computes the covariance matrix $S$, while Obs never does, now explains their names.) Below, we broadly describe how Cov and Obs parallelize CONCORD's bottlenecks; then we discuss the details.

**Parallelizing CONCORD's key bottlenecks.** *The Cov variant.* As mentioned, Cov proceeds by computing $S = \frac{1}{n}X^TX$ (line 2 in Algorithm 1), *i.e.*, the product of two dense matrices, upfront. Cov then uses $S$ to compute $W^{(k)} = \Omega^{(k)}S$, a sparse-dense product, on every line search iteration. The transpose $(W^{(k)})^T$ is formed on every proximal gradient iteration. All the matrix products can be computed using the communication-avoiding algorithm for dense-dense and sparse-dense matrix multiplication that we present below, while the transpose can be computed via partial all-to-all communication. Using $W^{(k)}$, Cov then computes the rest of the gradient $G^{(k)}$ (line 4) in an embarrassingly parallel way, as inverting a diagonal matrix (equivalent to inverting the entries on the diagonal) and applying the soft-thresholding operator $\mathcal{S}_{\tau\lambda_1}$ are just simple elementwise operations. As for the line search (lines 6–9): since (i) $\log \det(A) = \sum_i \log(A_{ii})$, (ii) $\mathbf{tr}(BC) = \sum_{i,j} B_{ij} C_{ij}$ and (iii) $\|B\|_F^2 = \mathbf{tr}(B^2)$, for a diagonal matrix $A$ and symmetric matrices $B, C$, these may be done in an embarrassingly parallel way.

*The Obs variant.* Obs never explicitly computes $S$; rather, it proceeds by computing $Y^{(k)} = \frac{1}{n}\Omega^{(k)} X^T$ on every proximal gradient and line search iteration via a communication-avoiding algorithm, as with Cov. The rest of the gradient $G^{(k)}$ is then computed by forming $Z^{(k)} = Y^{(k)} X$ as well as $(Z^{(k)})^T$, and using the same embarrassingly parallel elementwise operations as before. Noticing $\mathbf{tr}(\Omega^{(k)} S \Omega^{(k)}) = \frac{1}{n}\mathbf{tr}(\Omega^{(k)} X^T X \Omega^{(k)}) = \frac{1}{n}\|\Omega^{(k)} X^T\|_F^2 = \frac{1}{n}\|Y^{(k)}\|_F^2$, the line search (except for forming $Y^{(k)}$) also consists of elementwise operations.

We present a high-level description of the Cov variant of HP-CONCORD in Algorithm 2, highlighting the main differences vs. Algorithm 1 in blue. The pseudocode for Obs is presented in Algorithm 3.

**Matrix layouts and multiplication algorithms.** Parallel matrix multiplication is a well-studied problem, but most existing work considers dense matrices, and there is room for improvement in special cases. The most popular algorithm uses a 2D layout [2, 49], treating the processors as a square grid and making each processor responsible for all computations associated with one submatrix of the output matrix. 3D [1] and 2.5D [46] algorithms are provably communication-optimal in certain cases, and instead divide up the 3D iteration space, by essentially making $c$ copies of the output matrix, and having a square group of processors responsible for a subset of updates to that copy; the copies are eventually summed to produce the final answer. As we discuss next, these are not always the fastest methods in our setting, due to the matrix sizes and sparsity levels that arise with Cov and Obs.

*Cov.* For the sparse-dense product $W^{(k)} = \Omega^{(k)} S$, shifting around only $\Omega^{(k)}$ can use much less bandwidth, and could outperform the classic 2D/2.5D/3D algorithms by up to two orders of magnitude [29]. Therefore, we put $\Omega^{(k)}$ in 1D block row and $S$ in 1D block column layout. As for $S = \frac{1}{n} X^T X$: when $p > n$, $X^T$ is tall and $X$ is wide, so partitioning $X$ in a 2D layout, as 2D/2.5D/3D algorithms do, would result in tall and short local matrices, which perform poorly on local memory hierarchies. Instead, we group processors into teams of $c$ members each, and arrange the teams as a 1D array, distributing the rows of $X^T$ (a 1D block row layout) and the cols. of $X$ (a 1D block col. layout).

*Obs.* The dense-dense product $Z^{(k)} = Y^{(k)} X$ is similar to $X^T X$. For the sparse-dense product $Y^{(k)} = \Omega^{(k)} X^T$, we proceed just as for $W^{(k)} = \Omega^{(k)} S$ with Cov, putting $Y^{(k)}, \Omega, X^T$ all in a 1D block row layout.

Figure 1 illustrates all the distributed operations.

**Algorithm 2** The Cov variant of HP-CONCORD, for computing a sparse inverse covariance estimate.

**Input:** data matrix $X \in \mathbf{R}^{n \times p}$; tuning parameters $\lambda_1, \lambda_2 \geq 0$; optimization tolerance $\epsilon > 0$
**Output:** estimate $\hat{\Omega} \in \mathbf{R}^{p \times p}$ of the underlying inverse covariance matrix $\Omega^0$

1: $\Omega^{(0)} \leftarrow I$
2: Compute $S \leftarrow \frac{1}{n} X^T X$ ▷ Compute (once) via a distributed dense-dense matrix multiplication
3: Compute $W^{(0)} \leftarrow \Omega^{(0)} S$ ▷ Compute via a distributed sparse-dense matrix multiplication
4: **for** $k = 0, 1, 2, \ldots$
5:    Form $(W^{(k)})^T$ ▷ Form via a distributed matrix transpose
6:    $G^{(k)} \leftarrow -(\Omega_D^{(k)})^{-1} + \frac{1}{2}((W^{(k)})^T + W^{(k)}) + \lambda_2 \Omega^{(k)}$ ▷ Use $W^{(k)}, (W^{(k)})^T$
7:    $g(\Omega^{(k)}) \leftarrow -2 \sum_i \log(\Omega_{ii}^{(k)}) + \mathbf{tr}(W^{(k)} \Omega^{(k)}) + \frac{\lambda_2}{2} \|\Omega^{(k)}\|_F^2$ ▷ Use $(W^{(k)})^T$; see text for details
8:    **for** $\tau = 1, \frac{1}{2}, \frac{1}{4}, \ldots$
9:       $\Omega^{(k+1)} \leftarrow \mathcal{S}_{\tau \lambda_1}(\Omega^{(k)} - \tau G^{(k)})$ ▷ Apply the soft-thresholding operator, $\mathcal{S}_{\tau \lambda_1}$, in a distributed manner
10:       Compute $W^{(k+1)} \leftarrow \Omega^{(k+1)} S$ ▷ Compute via a distributed sparse-dense matrix multiplication
11:       $g(\Omega^{(k+1)}) \leftarrow -2 \sum_i \log(\Omega_{ii}^{(k+1)}) + \mathbf{tr}(W^{(k+1)} \Omega^{(k+1)}) + \frac{\lambda_2}{2} \|\Omega^{(k+1)}\|_F^2$ ▷ See text for details
12:    **until** $g(\Omega^{(k+1)}) \leq g(\Omega^{(k)}) - \mathbf{tr}((\Omega^{(k)} - \Omega^{(k+1)})^T G^{(k)}) + \frac{1}{2\tau} \|\Omega^{(k)} - \Omega^{(k+1)}\|_F^2$ ▷ See text for details
13: **until** a stopping criterion is satisfied, using $\epsilon$
14: **return** the estimate $\hat{\Omega} \leftarrow \Omega^{(k)}$

---

**Algorithm 3** The Obs variant of HP-CONCORD, for computing a sparse inverse covariance estimate.

**Input:** data matrix $X \in \mathbf{R}^{n \times p}$; tuning parameters $\lambda_1, \lambda_2 \geq 0$; optimization tolerance $\epsilon > 0$
**Output:** estimate $\hat{\Omega} \in \mathbf{R}^{p \times p}$ of the underlying inverse covariance matrix $\Omega^0$

1: $\Omega^{(0)} \leftarrow I$
2: Compute $Y^{(0)} \leftarrow \Omega^{(0)} X^T$ ▷ Compute via a distributed sparse-dense matrix multiplication
3: **for** $k = 0, 1, 2, \ldots$
4:    Compute $Z^{(k)} \leftarrow Y^{(k)} X$ ▷ Compute via a distributed dense-dense matrix multiplication
5:    Form $(Z^{(k)})^T$ ▷ Form via a distributed matrix transpose
6:    $G^{(k)} \leftarrow -(\Omega_D^{(k)})^{-1} + \frac{1}{2}((Z^{(k)})^T + Z^{(k)}) + \lambda_2 \Omega^{(k)}$ ▷ Use $Z^{(k)}, (Z^{(k)})^T$
7:    $g(\Omega^{(k)}) \leftarrow -2 \sum_i \log(\Omega_{ii}^{(k)}) + \frac{1}{n} \|Y^{(k)}\|_F^2 + \frac{\lambda_2}{2} \|\Omega^{(k)}\|_F^2$ ▷ Use $Y^{(k)}$; see text for details
8:    **for** $\tau = 1, \frac{1}{2}, \frac{1}{4}, \ldots$
9:       $\Omega^{(k+1)} \leftarrow \mathcal{S}_{\tau \lambda_1}(\Omega^{(k)} - \tau G^{(k)})$ ▷ Apply the soft-thresholding operator, $\mathcal{S}_{\tau \lambda_1}$
10:       Compute $Y^{(k+1)} \leftarrow \Omega^{(k+1)} X^T$ ▷ Compute via a distributed sparse-dense matrix multiplication
11:       $g(\Omega^{(k+1)}) \leftarrow -2 \sum_i \log(\Omega_{ii}^{(k+1)}) + \frac{1}{n} \|Y^{(k+1)}\|_F^2 + \frac{\lambda_2}{2} \|\Omega^{(k+1)}\|_F^2$ ▷ Use $Y^{(k+1)}$; see text for details
12:    **until** $g(\Omega^{(k+1)}) \leq g(\Omega^{(k)}) - \mathbf{tr}((\Omega^{(k)} - \Omega^{(k+1)})^T G^{(k)}) + \frac{1}{2\tau} \|\Omega^{(k)} - \Omega^{(k+1)}\|_F^2$ ▷ See text for details
13: **until** a stopping criterion is satisfied, using $\epsilon$
14: **return** the estimate $\hat{\Omega} \leftarrow \Omega^{(k)}$

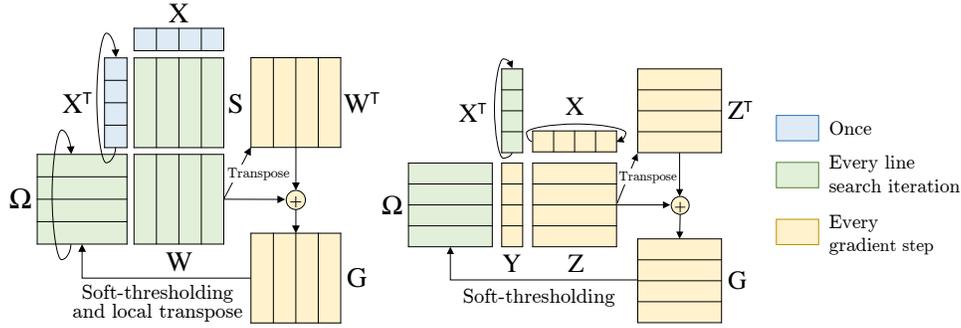

Figure 1: Left: Cov first computes $S = \frac{1}{n} X^T X$ by shifting $X^T$; then, on every iteration $k$, Cov computes $W^{(k)} = \Omega^{(k)} S$ by shifting $\Omega^{(k)}$, globally transposes $W^{(k)}$, computes $G^{(k)}$ from $W^{(k)}$ and $(W^{(k)})^T$, soft-thresholds $G^{(k)}$ to get $\Omega^{(k)}$, and converts $\Omega^{(k)}$ back to 1D block row layout by doing a local matrix transpose. Right: Obs computes $Y^{(k)} = \frac{1}{n} \Omega^{(k)} X^T$ by shifting $X^T$, computes $Z^{(k)} = Y^{(k)} X$ by shifting $X$, globally transposes $Z^{(k)}$ to get $(Z^{(k)})^T$ in the same layout, computes $G^{(k)}$ from $Z^{(k)}$ and $(Z^{(k)})^T$, and then soft-thresholds $G^{(k)}$ to get the new $\Omega^{(k)}$.

**Replication.** Here, we extend the 1.5D algorithm [29] to support different replication factors for each matrix operand. We use our algorithm, detailed in Algorithm 4, to compute all the matrix products arising with Cov and Obs. To compute the product $C = AB$, our algorithm rotates either $A$ or $B$ around (call this matrix $R$), fixing the other (call this $F$), and leaving $C$ stationary. Let $P$ be the number of homogeneous processors, let $c_R$ and $c_F$ be the replication factors of $R$ and $F$, and let $\mathbb{P}_R$ and $\mathbb{P}_F$ be the logical grids of processors of sizes $P/c_R \times c_R$ and $P/c_F \times c_F$, respectively. We partition $R$ equally, in 1D, into $P/c_R$ parts, and let processors $\mathbb{P}_R(i,:)$ have the $i$th part. Similarly, we partition $F$ and $C$ equally, in 1D, into $P/c_F$ parts, letting processors $\mathbb{P}_F(j,:)$ have the $j$th part. The processors $\mathbb{P}_F(j,:)$ now work together as a team to compute the $j$th part of $C$, with each member multiplying the $j$th part of $F$ with different parts of $R$. Further details are contained in Section S.2 of the supplementary material.

---

**Algorithm 4** Our 1.5D matrix multiplication algorithm.

1: **for each** $\mathbb{P}_R(i, \ell_R) = \mathbb{P}_F(j, \ell_F)$, in parallel, **do**
2:     $\delta \leftarrow \min(\ell_F, \ell_R) \cdot \max(1, c_F/c_R)$
3:     Shift $R$ by $\delta$
4:     **for** $\frac{p}{c_F c_R}$ rounds **do**
5:         Calculate local $C = AB$
6:         Shift $R$ by $c_F$
7:     **end for**
8:     SumReduce/Allgather $C$ between $\mathbb{P}_F(j,:)$
9: **end for**

---

**Final computation and communication costs.** We model the total running time of a distributed algorithm as $T = F\gamma + L\alpha + W\beta$, where $F$ is the total number of flops across all processors; $L$ is the *latency cost*, i.e., the total number of messages sent by all processors; and $W$ is the *bandwidth cost*, i.e., the size of all the messages sent. We define $\gamma, \alpha, \beta$ as machine-dependent quantities, measuring the time per flop, time to initiate a message, and time to send a *word* (i.e., an atomic unit of communication), respectively.

We also make the following definitions. Let $s$ be the total number of proximal gradient iterations for HP-CONCORD, $t$ be the average number of line search iterations on each proximal gradient iteration, and $d$ be the average number of nonzero entries in a row of $\Omega^{(k)}$ (averaged over all rows and all $st$ line search iterations). Also, let **nnz**$(C)$ denote the number of nonzeros in $C$.

The following lemma, proven by working through the details presented above for computing $W^{(k)} = \Omega^{(k)} S$, $S = \frac{1}{n} X^T X$, $Y^{(k)} = \Omega^{(k)} X^T$, and $Z^{(k)} = Y^{(k)} X$, presents the total dominant flop counts for Cov and Obs. Using the flop counts, the lemma also tells us when Cov is cheaper than Obs (when Cov is "worth it"). All our proofs are contained in Sections S.1 and S.2 of the supplementary material.

**Lemma 3.1.** *Cov costs $F_{Cov} = 2np^2 + 2dp^2(st+1)$ flops, while Obs costs $F_{Obs} = 2np^2 s + 2dnp(st+1)$ flops. Therefore, Cov is cheaper than Obs as long as:*

$$\frac{d}{p} < \frac{n}{p-n} \cdot \frac{1}{t}.$$

Turning now to the communication costs, our next lemma establishes that Algorithm 4 can reduce the communication involved in transposing a matrix, helpful as transposing can be an expensive operation since it involves all-to-all communication for a 1D layout.

**Lemma 3.2.** *In Algorithm 4, transposing the matrix $C$ requires: $\log_2(Q)$ messages and $(\mathbf{nnz}(C) \cdot c_R c_F \cdot Q \log_2(Q))/P$ words, where $Q = \max(P/c_R^2, P/c_F^2)$.*

The following lemma shows we can save a factor of $c_R c_F$ in latency and $c_R$ in bandwidth for each distributed matrix multiplication, by using Algorithm 4.

**Lemma 3.3.** *Algorithm 4 sends $\frac{P}{c_R c_F}$ messages and $\frac{\mathbf{nnz}(R)}{c_F}$ words, where $R$ is the matrix that is rotated.*

The next lemma presents the total communication costs for Cov and Obs, by using the preceding results and working through the communication costs of computing the various matrices.

**Lemma 3.4.** *Let $Q = \max(P/c_X^2, P/c_\Omega^2)$. Then Cov's communication costs are:*

$$L_{Cov} = \frac{P}{c_X^2} + st\frac{P}{c_X c_\Omega} + \log_2(Q),$$

$$W_{Cov} = \frac{np}{c_X} + st\frac{dp}{c_X} + p^2\frac{c_X c_\Omega}{P}Q\log_2(Q),$$

*while Obs' communication costs are:*

$$L_{Obs} = s(t+1)\frac{P}{c_\Omega c_X} + \log_2(Q),$$

$$W_{Obs} = s(t+1)\frac{np}{c_\Omega} + p^2\frac{c_X c_\Omega}{P}Q\log_2(Q).$$

Using the previously derived $F_{\text{Cov}}$, $L_{\text{Cov}}$, $W_{\text{Cov}}$, $F_{\text{Obs}}$, $L_{\text{Obs}}$, and $W_{\text{Obs}}$, our final lemma shows the total running times for Cov and Obs.

**Lemma 3.5.** *Cov's total running time is:*

$$T_{Cov} = [2np^2 + 2dp^2(st+1)]\gamma +$$
$$\left[\frac{P}{c_X^2} + st\frac{P}{c_X c_\Omega} + \log_2(Q)\right]\alpha +$$
$$\left[\frac{np}{c_X} + st\frac{dp}{c_X} + p^2\frac{c_X c_\Omega}{P}Q\log_2(Q)\right]\beta.$$

*while Obs' total running time is:*

$$T_{Obs} = [2np^2 s + 2dnp(st+1)]\gamma +$$
$$\left[s(t+1)\frac{P}{c_\Omega c_X} + \log_2(Q)\right]\alpha +$$
$$\left[s(t+1)\frac{np}{c_\Omega} + p^2\frac{c_X c_\Omega}{P}Q\log_2(Q)\right]\beta.$$

**Space complexity.** Asymptotically, both algorithms require at least $O(p^2)$ storage space: $O(c_X p^2)$ for Cov and $O(c_\Omega p^2)$ for Obs. In the future, we plan to reduce the space requirement by applying blocking with some recomputation. Currently, we simply scale up to more nodes when $p$ increases, since the computation complexity grows faster than the space complexity. As for the exact memory usage, both variants take the most space when computing $G$. Cov needs $S$, $\Omega$, $W$, and $W^T$ in memory ($G$ can be stored in place of $W$), Obs needs $\Omega$, $X$, $X^T$, $Y$, $Z$, and $Z^T$ in memory ($G$ can be stored in place of $Z$). Therefore, the memory requirements for Cov and Obs, respectively, are:

$$M_{\text{Cov}} = c_\Omega dp + 3c_X p^2,$$
$$M_{\text{Obs}} = 2c_X np + c_\Omega(dp + np + 2p^2).$$

Whether $T_{\text{Cov}}$ or $T_{\text{Obs}}$ is better (*i.e.*, lower) depends on the problem characteristics (*i.e.*, $n, p, d, s$, and $t$), the hardware parameters (*i.e.*, $\gamma$, $\alpha$, and $\beta$), and the replication factors (*i.e.*, $c_\Omega$ and $c_X$ subject to $c_\Omega c_X \leq P$, and $M_{\text{Cov}}, M_{\text{Obs}} \leq$ the amount of available memory).

## 4 Numerical examples

In this section, we evaluate the computational performance of HP-CONCORD on several synthetic data sets. We start by empirically verifying Lemma 3.1, running experiments with various values of $n$ and checking when Cov becomes faster than Obs. Afterwards, we investigate the benefits of replication, as discussed in Section 3, by varying the tuning parameters $c_X, c_\Omega$, controlling the amount of replication for the matrices $X, \Omega$, respectively. Finally, we run several head-to-head timing comparisons with BigQUIC.

As for the synthetic datasets, we consider both banded and random strictly diagonally dominant $\Omega^0$'s, corresponding to chain and random graphs, respectively, and sample Gaussian data. We fix the average degree to 2 for the chain graphs and 60 for the random graphs.

We implemented HP-CONCORD in C++ with OpenMP and MPI, and called threaded MKL for the local matrix multiplications. All our experiments were run with double precision on "Edison", a 5,586-node Cray XC30 supercomputer at the National Energy Research Scientific Computing Center (NERSC), where each node consists of two 12-core 2.4 GHz Intel Xeon E5-2695 processors with 64 GB of DDR3 RAM. We used 2 MPI processes per node.

**When does Cov become worth it?** For each graph type, we fix $p = 40{,}000$, vary $n \in \{100, 200, \ldots, 12{,}800\}$, and plot in Figure 2 the runtimes (on 16 nodes) required for Cov and Obs to generate estimates attaining the same average degree as the underlying graph. It is clear from the figure that Obs' runtime grows linearly with $n$, while Cov's does not, consistent with Lemma 3.1, as the dominant term in Obs' cost depends linearly on $n$ and Cov's does not. (Cov required more iterations to converge for $n = 100$ and $n = 200$ than for for larger values of $n$, and therefore the runtimes for $n = 100$ and $n = 200$ are higher: when $n = 100, 200, \ldots, 12{,}800$, Cov required 28, 21, 20, 18, 17, 16, 16, and 15 iterations for the chain graphs, and 155, 102, 47, 58, 58, 66, 61, and 76 iterations for the random graphs.) The trend reverses when we fix $n$ and increase $p$, *i.e.*, Obs becomes asymptotically more appealing than Cov. Interestingly, the crossover point where Cov becomes faster than Obs happens later than Lemma 3.1 predicts, since most of Cov's cost comes from sparse-dense matrix multiplications, which have higher time per flop than the dense-dense matrix multiplications that dominate Obs' overall cost (*i.e.*, $\gamma_{\text{sparse-dense}} \gg \gamma_{\text{dense-dense}}$).

**Benefits of replication.** To illustrate the benefits of replication, we run Obs on all possible replication configurations (*i.e.*, all $c_X, c_\Omega$ combinations) with 256 nodes on a chain graph, where $p = 40k$, $n = 100$, plotting all the runtimes in seconds in Figure 3. The figure shows both extremes:

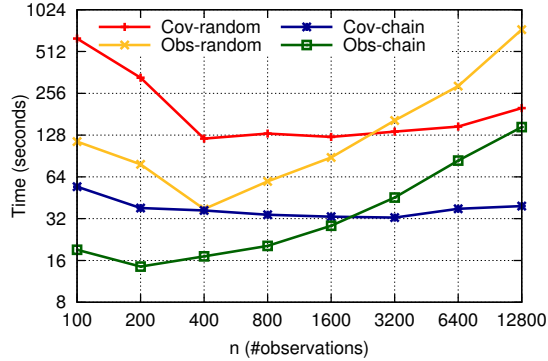

Figure 2: Runtimes in seconds for Cov (blue/red) vs. Obs (green/yellow), on synthetic data coming from chain and random graphs, with 16 nodes, $p = 40\text{k}$, and various $n$.

|  | $c_X$ | | | | | | | | |
|---|---|---|---|---|---|---|---|---|---|
| $c_\Omega$ | 1 | 2 | 4 | 8 | 16 | 32 | 64 | 128 | 256 | 512 |
| 1 | 13.15 | 9.69 | 8.73 | 7.34 | 7.09 | 7.01 | 7.17 | 6.56 | 7 | 5.01 |
| 2 | 9.35 | 5.75 | 5.78 | 7.1 | 5.42 | 5.42 | 5.78 | 5.13 | 3.9 | |
| 4 | 5.28 | 3.94 | 3.42 | 5.08 | 4.44 | 4.39 | 4.3 | 3.86 | | |
| 8 | 4.08 | 3.45 | 3.07 | 2.84 | 4.19 | 4.08 | 3.65 | | | |
| 16 | 4 | 3.55 | 3.2 | 2.63 | 3.83 | 3.77 | | | | |
| 32 | 4 | 6.06 | 3.43 | 3.38 | 3.26 | | | | | |
| 64 | 6.13 | 4.77 | 4.7 | 4.42 | | | | | | |
| 128 | 7.8 | 6.59 | 5.87 | | | | | | | |
| 256 | 10.97 | 9.47 | | | | | | | | |

Good → Bad

Figure 3: Runtimes in seconds for Obs, at various replication factors, on a chain graph with 256 nodes × 2 MPI processes per node, $p = 40\text{k}$, and $n = 100$. Colder/warmer cells mean better/ worse runtimes. The worst runtime is due to the non-communication-avoiding configuration, circled in blue; the best is circled in red, with a 5× speedup due to replication.

(a) when $c_X = c_\Omega = 1$, Obs is in a purely non-communication-avoiding mode, partitioning all the matrices into $P$ equal blocks, and takes the longest to run; and (b) when $c_X = 512$ and $c_\Omega = 1$, every processor maintains the entire $X$, does all matrix multiplications locally, and only communicates when replicating $X$ and during the transpose. The best result comes when $c_X = 8$ and $c_\Omega = 16$, a 5× speedup over the non-communication-avoiding configuration.

**Comparison with BigQUIC.** Finally, we compare HP-CONCORD with BigQUIC, on chain and random graphs. To put the two algorithms on an equal footing, we choose the tuning parameters so that the estimates are equally sparse. For the chain graphs, we fix $n = 100$ and vary $p$ from 10,000 to 1.28 million. As $d$ is not much smaller than $n$, we use the Obs variant of HP-CONCORD and vary the number of nodes, reporting the best runtime across a range of replication levels. We present the results in Figure 4a, and interpolate BigQUIC's result for $p = 1{,}280{,}000$ as it took longer than four days (the maximum running time allowed on Edison) to converge. From the figure, we see HP-CONCORD matching BigQUIC in a shared memory setup (*i.e.*, one node; BigQUIC cannot run on multiple nodes), and demonstrating good scalability as more nodes are added, which the user may customize to fit their needs; *e.g.*, when $p = 80,000$, an estimate may be computed in less than four seconds with 1,024 nodes.

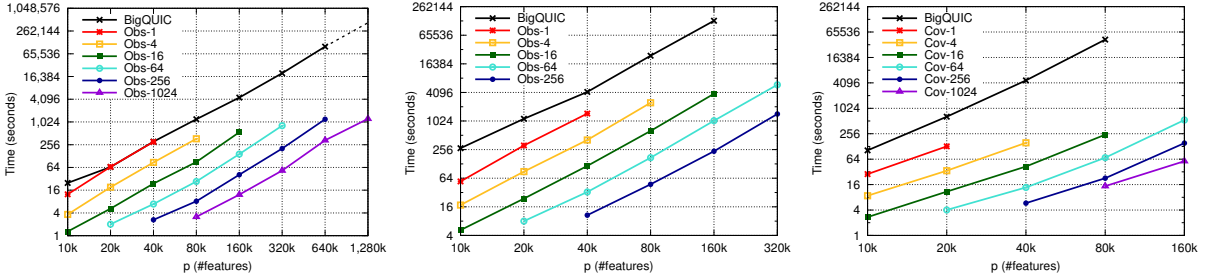

(a) Chain graph with $n = 100$.  (b) Random graph with $n = 100$.  (c) Random graph with $n = p/4$.

Figure 4: Runtimes in seconds for HP-CONCORD vs. BigQUIC on various graph types. "Obs-x" or "Cov-x" denotes Obs or Cov run with x nodes. By design, BigQUIC only runs on 1 node (24 cores).

For the random graphs, we vary $p$ from 10,000 to 320,000. In Figure 4b, we fix $n = 100$, and again use Obs, as $d$ here is even larger than it was for the chain graphs. In Figure 4c, we set $n = p/4$, and use Cov because $n$ is large. In both figures, we see Obs outperforming BigQUIC by about an order of magnitude, and exhibiting even better scalability as more nodes are added than with the chain graphs, since $\Omega^0$ is comparatively denser here. Lastly, in Table 1, we present the number of iterations required to converge as well as the best positive predictive values and false discovery rates (computed by looking at the differences between the estimated and true sparsity patterns, across a range of tuning parameters), for both methods. BigQUIC naturally takes fewer iterations to converge than HP-CONCORD, as BigQUIC is a second-order method. On the other hand, at all problem sizes, our method demonstrates better graph recovery than BigQUIC.

| Graph type | Method | $p = 10k$ | $p = 20k$ | $p = 40k$ | $p = 80k$ | $p = 160k$ | $p = 320k$ | $p = 640k$ | $p = 1,280k$ |
|---|---|---|---|---|---|---|---|---|---|
| Chain | BigQUIC | 6 | 5 | 6 | 6 | 5 | 5 | 5 | – |
| ($n = 100$) | HP-CONCORD | 25 | 33 | 37 | 36 | 43 | 51 | 69 | 57 |
| Random | BigQUIC | 6 | 6 | 5 | 6 | 5 | – | – | – |
| ($n = 100$) | HP-CONCORD | 114 | 144 | 155 | 203 | 270 | 330 | – | – |
| | BigQUIC | 5 | 5 | 5 | – | – | – | – | – |
| | • PPV | 99.48 | 99.71 | 99.78 | 99.81 | – | – | – | – |
| | • FDR | 0.52 | 0.29 | 0.22 | 0.19 | – | – | – | – |
| Random | HP–CONCORD | 16 | 17 | 17 | 21 | 35 | – | – | – |
| ($n = p/4$) | • PPV | 99.75 | 99.92 | 99.94 | 99.94 | 99.20 | – | – | – |
| | • FDR | 0.25 | 0.08 | 0.06 | 0.06 | 0.80 | – | – | – |

Table 1: Each cell indicates the number of iterations required for BigQUIC and HP-CONCORD to converge in a particular chain or random graph experiment. Also indicated, for the random graph experiments with $n = p/4$, are the positive predictive values (PPVs) and false discovery rates (FDRs) for HP-CONCORD and BigQUIC, relative to the sparsity pattern of $\Omega^0$, for various $p$; these numbers are expressed as percentages. "–" indicates that a method took longer than four days to converge.

# 5 Case study: graph estimation from fMRI data

We now present a case study, where we use HP-CONCORD to make progress on a challenging and important problem in neuroscience: obtaining a biologically meaningful clustering of the brain (more specifically, points on the cerebral cortex), from high-dimensional fMRI data. The data we use, from the Human Connectome Project [45], is a $(91,282 \times 91,282)$-dimensional sample covariance matrix, roughly 60 GB in size, and requires fitting $\approx$4 billion parameters. The large number of dimensions rules out most methods for sparse inverse covariance estimation, but makes HP-CONCORD a natural choice. Below, we qualitatively and quantitatively compare the clusterings generated by HP-CONCORD to those from Glasser et al. [24], a state-of-the-art clustering from the neuroscience literature, presented in Figure 5. Previewing our findings, we see that the entirely data-driven clusterings generated by HP-CONCORD are able to capture many of the important features also present in Glasser et al.

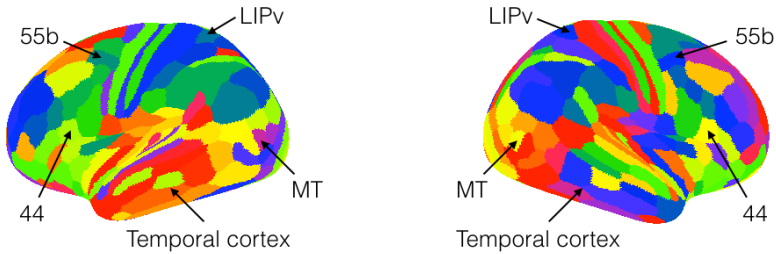

Figure 5: The clusterings from Glasser et al., for the left and right hemispheres of the brain; the colors only serve to differentiate the different clusters and carry no additional meaning. The clusterings rely on multiple exogenous data sources and a significant amount of domain knowledge.

**Approach.** We pursue a two-step approach, where for step (i) we generate a partial correlation graph using HP-CONCORD, and for step (ii) we apply a graph-based clustering algorithm to the partial correlation graph arising from the sparsity pattern of the HP-CONCORD estimate. For step (i), we consider all combinations of the tuning parameters $\lambda_1 \in\{$0.48, 0.5, 0.52, 0.54, 0.57, 0.59, 0.61, 0.64, 0.67, 0.69, 0.72$\}\times\lambda_2 \in\{$0.10, 0.13, 0.16, 0.2, 0.25, 0.31, 0.39, 0.49$\}$; tuning parameters outside these ranges yielded either trivially sparse or dense estimates. Running HP-CONCORD on a single $(\lambda_1, \lambda_2)$ pair took $\approx$37 minutes on Edison. For step (ii), the clustering algorithms we consider are: the well-known Louvain method [13], as well as a relatively new clustering method from the persistent homology literature [21] that leverages the degree matrix associated with the partial correlation graph (details in Section S.3.4 of the supplementary material). Additionally, because the clusterings from Glasser et al. treat the left and right hemispheres of the brain separately, we run and evaluate our clustering algorithms on the subgraphs for the left and right hemispheres separately. We ran our code on "Eos", a 736-node Cray XC30 supercomputer with 16-core 2.6 GHz Intel Xeon E5-2670 at Oak Ridge Leadership Computing Facility (OLCF).

**Evaluation.** As mentioned, our main points of comparison are the recent clusterings, for the left and right hemispheres, from Glasser et al., presented in Figure 5. However, we also consider a simple baseline, given by discarding $\{$99, 99.1, ..., 99.8, 99.9, 99.91, ..., 99.98, 99.99$\}$% of the sample covariance matrix entries: *i.e.*, keep entries with the largest magnitudes (c.f. [35]) in order to generate marginal correlation graphs. This baseline lets us probe the comparative advantage of using marginal vs. partial correlations. To quantitatively compare clusterings, we consider a variant of the Jaccard score; details are in Section S.3.5 of the supplementary material.

**Results.** The top and middle rows of Table 2 present the best clusterings generated by the Cov variant of HP-CONCORD, followed by the persistent homology and Louvain methods, respectively, when compared to those of Glasser et al. presented in Figure 5, according to the (modified) Jaccard score. The bottom row presents the best clusterings generated by thresholding the sample covariance matrix at various levels. The left and middle columns present the results for the left and right hemispheres, respectively. We see that the persistent homology clusterings perform the best, in terms of Jaccard score, across both hemispheres.

Qualitatively, the persistent homology clusterings are able to identify several clusters of interest to the neuroscience community (c.f. Figure 3 in [24]); this is certainly encouraging, since we do not expect perfect recovery of all the clusters in Figure 5, as the latter clusters rely on multiple exogenous data sources and a significant amount of domain knowledge. Some examples: the persistent homology clusterings seem to pick out area 55b, involved in hearing; the lateral intraparietal cortex (LIPv), involved in eye movement; and much of the variation in the temporal cortex, involved in processing information from the senses. On the other hand, the Louvain method and the clusterings generated by the sample covariance matrix seem to miss these clusters, as they appear overly smooth. All the methods seem to miss Brodmann's area 44, involved in hearing and speaking, and the middle temporal visual area (MT), involved in seeing moving objects.

Lastly, it is also interesting to analyze the sparsity patterns of the HP-CONCORD estimates; details are contained in Section S.3.3 of the supplementary material.

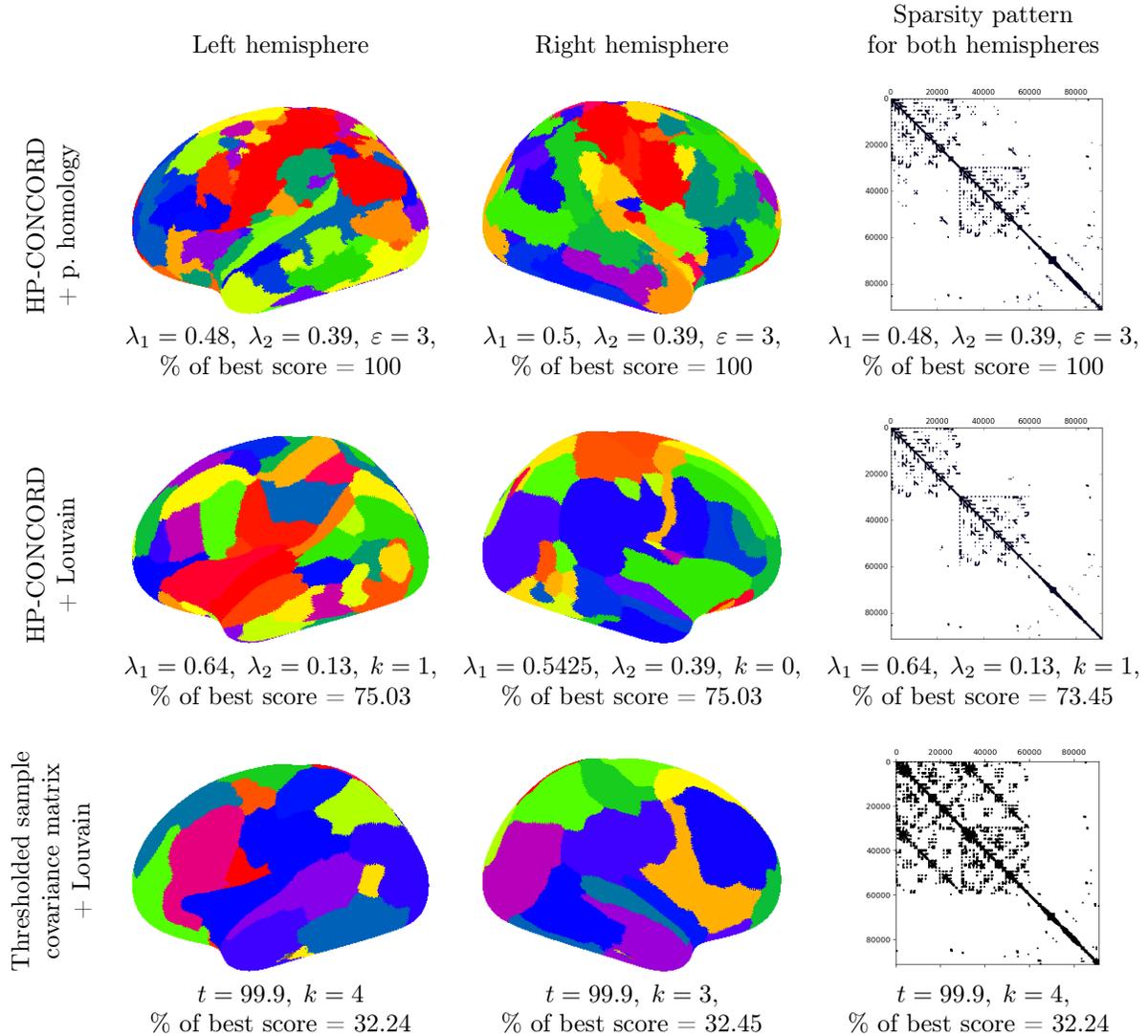

Table 2: The left and middle columns present the best clusterings for the left and right hemispheres, respectively, according to the (modified) Jaccard score, relative to those of Glasser et al.; the colors only serve to differentiate the different clusters. The right-most column presents the sparsity patterns for both hemispheres yielding the best clustering for the left hemisphere; black indicates a nonzero entry. The rows correspond to the various methods. The scores under the figures are the percentages of the best Jaccard score attained; higher is better. Since the persistent homology clusterings perform the best, these percentages are just 100. Also indicated are the tuning parameter values $(\lambda_1, \lambda_2, \varepsilon, k, t)$ yielding the best clusterings. The Jaccard scores, tuning parameter details, and an expanded set of results are in Section S.3 of the supplementary material.

# 6  Conclusion

We presented HP-CONCORD, a communication-avoiding distributed proximal gradient method for estimating a sparse inverse covariance matrix from "massive-scale" data. HP-CONCORD can fit ≈819 billion parameters ($p = 1.28$ million) in ≈17 minutes in a distributed memory setting, and is an order of magnitude faster than BigQUIC when fitting ≈800 million parameters ($p = 40$k) in a shared memory setting. For future work, it may be interesting to apply a divide-and-conquer strategy based on a block structure assumption, to further optimize the computation. We also used HP-CONCORD to capture the functional connectivity structure of the cerebral cortex as a partial correlation graph, which, in turn, was clustered into distinct regions; some regions showed good agreement with a result from the neuroscience literature.

## Acknowledgements

This work was supported in part by: the Department of Energy's Office of Science Advanced Scientific Computing Research (DOE/SC/ASCR) X-Stack and Applied Mathematics programs at LBNL (DEAC02-05CH11231) and at UC Berkeley (UCB) (DE-SC0008700); the Department of Defense; the Gordon and Betty Moore Foundation at UCB (GBMF3834); the Alfred P. Sloan Foundation at UCB (2013-10-27); a DOE Computational Science Graduate Fellowship (DE-FG02-97ER25308); and by UCSB, including a Faculty Research Grant. It used resources supported by DOE/SC/ASCR at the National Energy Research Scientific Computing Center (DE-AC02-05CH11231) and the Oak Ridge Leadership Computing Facility (DE-AC05-00OR22725). The U.S. Government (USG) retains, and the publisher, by accepting the article for publication, acknowledges, that the USG retains a non-exclusive, paid-up, irrevocable, world-wide license to publish or reproduce the published form of this manuscript, or allow others to do so, for USG purposes. We thank Ryan J. Tibshirani and J. Zico Kolter for the helpful feedback.

# Supplementary Material for the Paper "Communication-Avoiding Optimization Methods for Distributed Massive-Scale Sparse Inverse Covariance Estimation"

The supplementary material that follows contains proofs and additional details for the paper "Communication-Avoiding Optimization Methods for Distributed Massive-Scale Sparse Inverse Covariance Estimation"; all section, equation, table, and figure numbers are preceded by the letter "S" (all section, equation, table, and figure numbers without an "S" refer to the main paper).

## S.1 Computational cost comparison for Cov vs. Obs

Cov is preferred over Obs when the flop count for Cov is less than that for Obs (from Lemma 3.1 in the main paper), *i.e.*,

$$
\begin{aligned}
2np^2 + 2dp^2(st+1) &< 2np^2 s + 2dnp(st+1) \\
2dp(st+1)(p-n) &< 2np^2(s-1) \\
d(st+1)(p-n) &< np(s-1).
\end{aligned}
\tag{S.1}
$$

We relax the comparison a little by plugging in $st < st+1$ and $s > s-1$:

$$
\begin{aligned}
dst(p-n) &< nps \\
\frac{d}{p} &< \frac{n}{p-n} \cdot \frac{1}{t}.
\end{aligned}
\tag{S.2}
$$

Let $r_{\text{obs}} = n/p$ be the ratio of the number of observations to the number of features and let $r_{\text{nnz}} = dp/p^2 = d/p$ be the *average* percent nonzeroes of $\Omega$ throughout all iterations, $0 < r_{\text{obs}}, r_{\text{nnz}} \leq 1$, we can reformulate (S.2) as

$$r_{\text{nnz}} < \frac{r_{\text{obs}}}{1 - r_{\text{obs}}} \cdot \frac{1}{t}.$$

Above, $r_{\text{obs}}/(1 - r_{\text{obs}})$ is an increasing function. The closer $n$ is to $p$, the higher $r_{\text{nnz}}$ can be for Cov to still require less flops than Obs. For example, assume $t = 10$ (we observed 5-15 line search iterations per one proximal gradient iteration in practice). Substituting $r_{\text{obs}} = 0.01, 0.1$, and $0.25$ give $r_{\text{nnz}} < 0.001, 0.011$, and $0.033$, respectively. Applications with *average* percent nonzeroes of $\Omega$ (throughout all iterations) less than $0.1\%, 1.1\%$, and $3.3\%$ should benefit from Cov in these cases.

## S.2 Additional details on our 1.5D matrix multiplication algorithm

This subsection details how Algorithm 4 in the main paper computes $C = AB$. Our algorithm partitions all matrices in a 1D layout. We pick one operand to move around ($A$ or $B$) and fixing the other operand and $C$ stationary. We will call the rotating matrix $R$ and the other fixed operand $F$. The notation we use is summarized in the table below. Our indexing is one-based and cyclic.

| Notation | Meaning |
|---|---|
| $P$ | Total number of processors |
| $R$ | Matrix operand being rotated ($A$ or $B$) |
| $F$ | Matrix operand being fixed ($B$ or $A$) |
| $c_R$ | Replication factor of $R$ |
| $c_F$ | Replication factor of $F$ and $C$ |
| $\mathbb{P}_R$ | Logical grid of processors of size $P/c_R$ teams by $c_R$ layers |
| $\mathbb{P}_F$ | Logical grid of processors of size $P/c_F$ teams by $c_F$ layers |

### S.2.1 Data layout

We partition $R$ in 1D into $P/c_R$ equal parts: $R(1), \ldots, R(P/c_R)$ and let processors $\mathbb{P}_R(i, :)$ own $R(i)$. There are $c_R$ processors in each team $\mathbb{P}_R(i, :)$ so each $R(i)$ is stored $c_R$ times across all processors, hence why we call $c_R$ the replication factor of $R$. Likewise, we partition $F$ and $C$ in 1D into $P/c_F$ equal parts: $F(1), \ldots, F(P/c_F)$ and $C(1), \ldots, C(P/c_F)$. We let processors $\mathbb{P}_F(j, :)$ own $F(j)$ and $C(j)$.

### S.2.2 Multiplication

Team $\mathbb{P}_F(j, :)$ is responsible for computing $C(j)$ together. Each team member computes appropriate multiplication between the fixed matrix $F(j)$ with different parts of $R$. For convenience, we reproduce the pseudocode for Algorithm 4 from the main paper below.

---
**Algorithm 4** Our 1.5D matrix multiplication algorithm.
---
1: **for each** $\mathbb{P}_R(i, \ell_R) = \mathbb{P}_F(j, \ell_F)$, in parallel, **do**
2:    $\delta \leftarrow \min(\ell_F, \ell_R) \cdot \max(1, c_F/c_R)$
3:    Shift $R$ by $\delta$.
4:    **for** $\frac{p}{c_F c_R}$ rounds **do**
5:       Calculate local $C = AB$
6:       Shift $R$ by $c_F$
7:    **end for**
8:    SumReduce/Allgather $C$ between $\mathbb{P}_F(j, :)$
9: **end for**
---

Line 2 calculates an offset $\delta$ that each member would shift $R$ initially to all get different parts of $R$. Line 3 performs the initial shift. Then we alternate between multiplying the local matrices and shifting $R$ by $c_F$ to get a new part of $R$. There are $P/c_R$ different parts of $R$ and each team calculates $c_F$ parts at once (one part per each team member). The algorithm needs to run for $P/(c_R c_F)$ rounds.

Figure S.1 shows an example when we partition $A$ in 1D block row layout, $B$ and $C$ in 1D block column layout, and rotate $A$, on 16 processors. Figure S.1a shows the situation when $c_A \leq c_B$. The

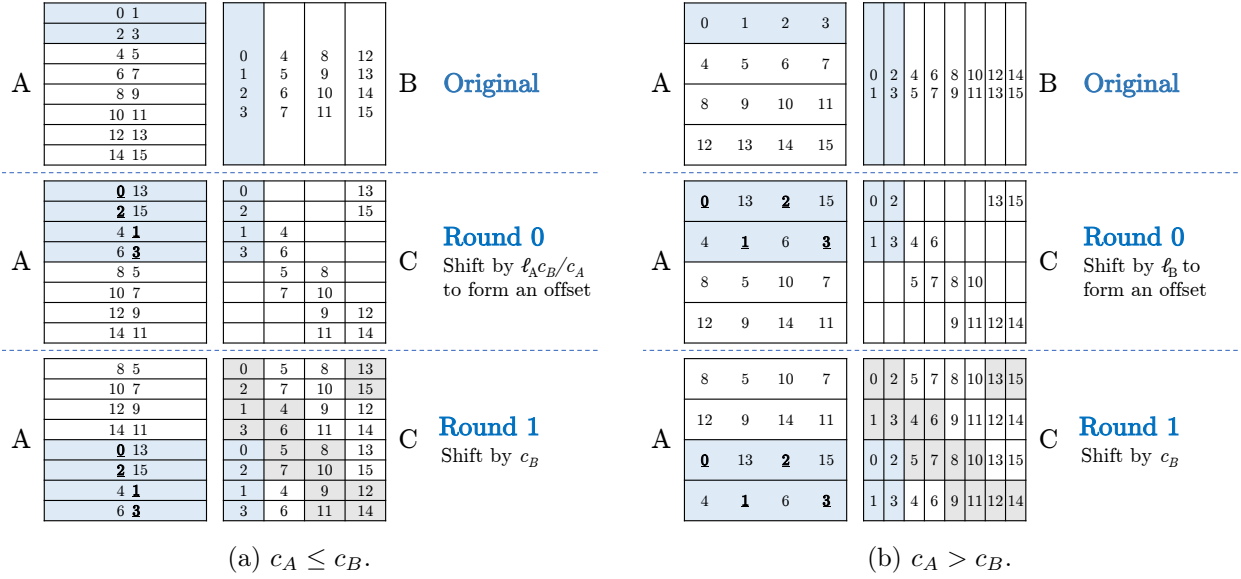

Figure S.1: Processors $\mathbb{P}_B(j,:)$ compute $C(j)$ together. The numbers on the matrix parts are the ranks of the processors that it resides on. Here, $p = 16$ and $(c_A, c_B)$ is $(2, 4)$ in (a) and $(4, 2)$ in (b). The first line shows the original layouts of $A$ and $B$. The second line (Round 0) shifts $A$ by $\delta$ to compute the first $c_B$ blocks of $C(j)$. The third line (Round 1) shifts $A$ by $c_B$ and computes the rest of $C$.

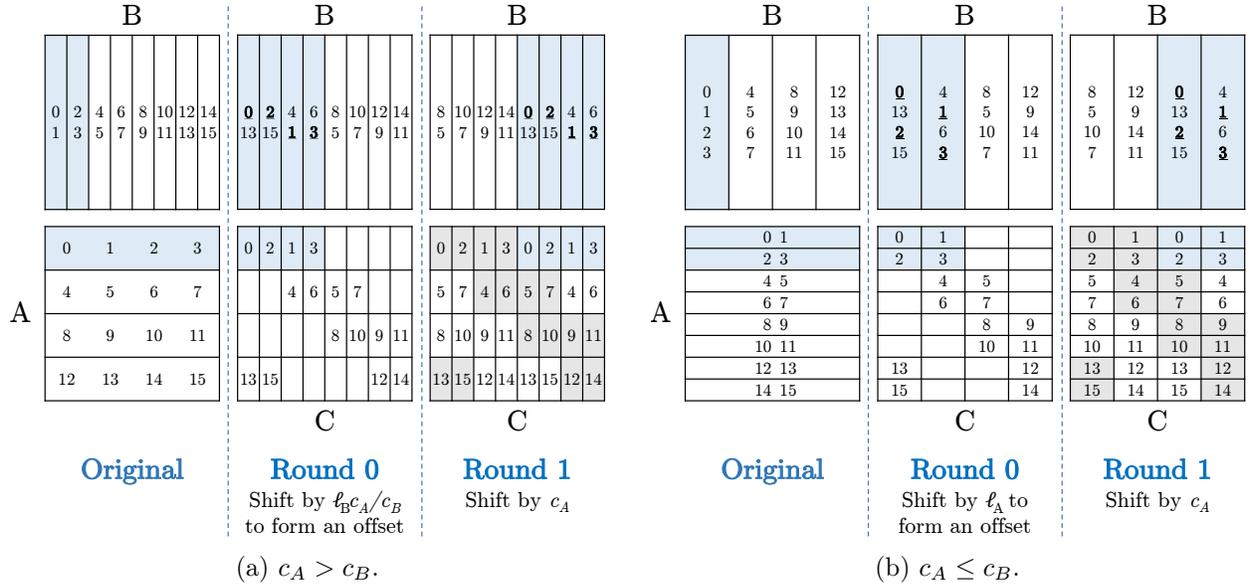

Figure S.2: Processors $\mathbb{P}_A(i,:)$ compute $C(i)$ together. The numbers on the matrix parts are the ranks of the processors that it resides on. Here, $p = 16$ and $(c_A, c_B)$ is $(4, 2)$ in (a) and $(2, 4)$ in (b). The Original column shows the original layouts of $A$ and $B$. Round 0 shifts $B$ by $\delta$ to compute the first $c_A$ blocks of $C(i,:)$. Round 1 shifts $B$ by $c_A$ and computes the rest of $C$.

first row shows the original layout of $A$ and $B$. Written on each matrix part are the processor ranks. $c_A = 4$ and $c_B = 2$. The second row shows Round 0, where each processor performs an initial shift so that $\mathbb{P}_B(j,:)$ as a whole would multiply $A(jc_B/c_A)$ to $A(jc_B/c_A + c_B)$ with $B(j)$. The next round the team moves to the next $c_B$ blocks of $A$ and finish the rest of the multiplications. Figure S.1b shows what happens when $c_A > c_B$. Figure S.2 illustrates how the algorithm works when we shift $B$ instead of $A$.

### S.2.3 Communication costs

Let $\mathbf{nnz}(\cdot)$ denote the number of nonzeroes of a matrix, sparse or dense. There are $p/(c_R c_F)$ rounds, each round each processor sends a message. The message size is $\mathbf{nnz}(R)/(P/c_R) = c_R \mathbf{nnz}(R)/P$. The total number of messages and words are

$$L_{1.5\mathrm{D}} = \frac{P}{c_R c_F} \text{ messages,}$$

$$W_{1.5\mathrm{D}} = \frac{P}{c_R c_F} \cdot \frac{c_R \mathbf{nnz}(R)}{P} = \frac{\mathbf{nnz}(R)}{c_F} \text{ words.}$$

### S.2.4 Transposing the resulting matrix $C$

Normal 1D partitioning without replication requires all $P$ processors to talk to all other processors in a transpose. Replication helps us limit the number of processors each processor has to exchange matrices with. For example, in Figure S.1a, processor 0 has to exchange sub-matrices with processors 2, 8, and 10 to do a transpose, not all processors anymore. For each team $\mathbb{P}_F(j,:)$, there are $P/c_R$ block rows of $C(j)$ to calculate, so each processor $\mathbb{P}_F(j, \ell_F)$ calculates $P/(c_R c_F)$ submatrices of $C(j)$. When being transposed, each submatrix will span $\max(c_R/c_F, c_F/c_R)$ submatrices from other teams. The number of processors each processor has to communicate to in a transpose is

$$\frac{P}{c_R c_F} \max\left(\frac{c_R}{c_F}, \frac{c_F}{c_R}\right) = \max\left(\frac{P}{c_F^2}, \frac{P}{c_R^2}\right).$$

Each processor holds $\mathbf{nnz}(C)/(P/c_F)$ words of $C$, split into $P/c_R$ blocks. Therefore, the number of words each block to be transposed has is

$$\mathbf{nnz}(C) \cdot \frac{c_R c_F}{P^2}.$$

An all-to-all communication between $P$ processors where everyone sends and receives $w$ words from everyone takes $O(\log P)$ messages and $O(wP \log P)$ words. Therefore, the costs of transposing the resulting matrix $C$ from Algorithm 4 are

$$L_{\mathrm{xpose}} = O\left(\log \max\left(\frac{P}{c_F^2}, \frac{P}{c_R^2}\right)\right) \text{ messages,}$$

$$W_{\mathrm{xpose}} = O\left(\frac{\mathbf{nnz}(C) c_R c_F}{P^2} \max\left(\frac{P}{c_F^2}, \frac{P}{c_R^2}\right) \log \max\left(\frac{P}{c_F^2}, \frac{P}{c_R^2}\right)\right) \text{ words.}$$

## S.3 Additional details, figures, and tables for the fMRI experiments

### S.3.1 Further details on the fMRI data

Here, we present further details on the fMRI data, and in particular the sample covariance matrix, we use. The sample covariance matrix we use was generated in the following way (c.f. Figure 2 of [45]). First, 1,200 subjects were put into a state-of-the-art fMRI machine and measurements were taken without stimulating the subjects, every 0.7 seconds for an hour, at 2 millimeter $\times$ 2 millimeter $\times$ 2 millimeter cubes/voxels spread evenly throughout the cerebral cortex. Next, as fMRI data is typically very noisy, a significant amount of post-processing was done to denoise the data, ultimately leading to a data matrix $X$ with dimensions $n \approx 6{,}171{,}400$, $p = 91{,}282$. To further reduce the level of noise, the columns of the data matrix were then averaged over the 1,200 subjects, leading to a data matrix with dimensions $n \approx 5{,}142$, $p = 91{,}282$, from which the sample covariance matrix was finally computed.

### S.3.2 Related work

We mention some work that is related to our approach, noting that most other works (i) do not use the latest available fMRI data, (ii) are not particularly scalable, and/or (iii) use nonconvex estimators, which can lead to interpretability issues. The work of [44] applies the "SPACE" estimator of [40], except augmented with an additional squared $\ell_2$-norm penalty, to low-dimensional fMRI data; while certainly related to our approach, this method unfortunately suffers from all three of the aforementioned issues. The work of [23] applies the graphical lasso algorithm of [22] to older fMRI data, except with an $\ell_0$-"norm" penalty instead of the usual $\ell_1$-norm penalty, making the corresponding optimization problem nonconvex. The work of [11] applies a pseudolikelihood-based estimator with a deep neural network as the predictive primitive (instead of, say, a lasso regression) to fMRI data obtained from subjects with autism instead of the broader population. Lastly, the works of [41, 42] focus on scalable methods for structure recovery from fMRI data, rather than for the broader problem of sparse inverse covariance estimation.

### S.3.3 Further details on the sparsity patterns of the HP-CONCORD estimates

It is interesting to analyze the sparsity patterns of the HP-CONCORD and sample covariance matrix estimates yielding the best clusterings; these are presented in the left-most column of Table 2. Three features here are striking: (i) the HP-CONCORD estimates possess a block diagonal structure, where the blocks turn out to correspond to the left and right hemispheres; (ii) furthermore, the sparsity patterns of the blocks themselves turn out to correspond to the (spatially) closest voxels; (iii) although the sparsity patterns of the HP-CONCORD and sample covariance matrix estimates appear vaguely similar, the subtle differences between them drive the (significant) differences in the clusterings. We emphasize that these features arise naturally, without being hard-coded into our method.

The sparsity patterns of the diagonal blocks of the HP-CONCORD estimates turn out to correspond to the (spatially) closest voxels on the left and right hemispheres; here we provide some further details. In Figure S.3, we present the sparsity pattern of the HP-CONCORD estimate attaining the best clustering, for the left hemisphere, in the left column of Table 2 in the main paper; in Figure S.3, we also present the sparsity pattern of a matrix we constructed, where the $(i,j)$th entry of the matrix is the great-circle distance between the voxels $i$ and $j$, after retaining only 0.1% of closest voxels. The sparsity patterns of the distance matrix and HP-CONCORD estimates indeed look visually similar, suggesting that the best HP-CONCORD estimate has recovered some of the spatial signal in the data, without being "told" to do so. Inspecting the right hemisphere conveys the same message.

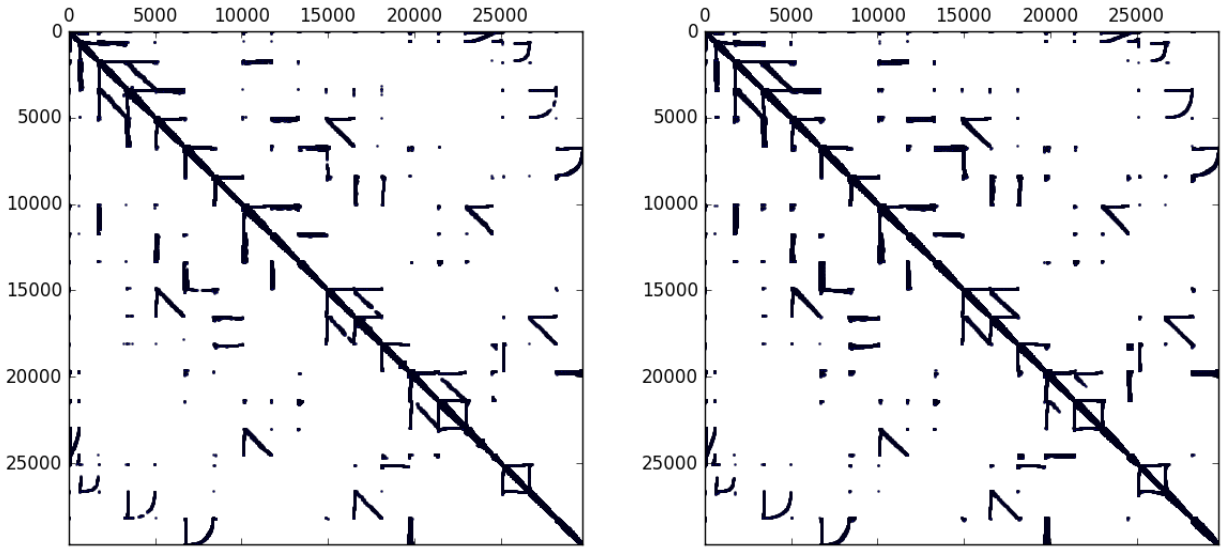

Figure S.3: Left: the sparsity pattern of the HP-CONCORD estimate attaining the best clustering in the left column of Table 2 in the main paper, where we have only plotted the 29,696 coordinates belonging to the left hemisphere. Right: the sparsity pattern of a $(91,282 \times 91,282)$-dimensional (symmetric) matrix we constructed, whose $(i,j)$th entry is the great-circle distance between the voxels $i$ and $j$, where we have (again) only plotted the 29,696 coordinates belonging to the left hemisphere, and additionally retained just the 0.1% of closest voxels. In both plots, black indicates a nonzero entry.

### S.3.4 Details on persistent homology-based clustering algorithm

We map the degree of a vertex in the inverse covariance graph, described by matrix $\Omega^{p \times p}$, onto the surface of a brain. We thus obtain a function $f : S \to \mathbb{Z}$, where $S$ is the triangulation of the cortical surface. We apply the watershed algorithm [15] to $f$ by sweeping the vertices from the highest value to the lowest. We start a new label if the vertex has no labeled neighbors in $S$. If it does, we propagate the label with the maximum starting value.

The resulting parcellation is usually too fine: every local maximum of $f$ produces a new label. We use the theory of persistent homology [21] to coarsen the parcellation. During the sweep, we build the dual graph $G$ of the labels. When we start a new label $l$ at a vertex $u$, we add $l$ to the graph and assign to it the value $f(u)$. When two labels, $l_1$ and $l_2$, that fall in different components of $G$, meet at a vertex $v$ during the sweep, we add an edge $(l_1, l_2)$ to $G$. We find the maximum values, $a_1$ and $a_2$, assigned to any vertex in the components of $l_1$ and $l_2$ in $G$, and assign to the new edge the value $\min\{a_1, a_2\} - f(v)$. (It's not difficult to verify that this is exactly the persistence of the vertex $v$.)

Once we construct the dual graph $G$, given a simplification threshold $\varepsilon$, we treat the connected components of the subgraph of $G$ induced by the edges with values at most $\varepsilon$ as the new parcels. As we increase $\varepsilon$, the parcels merge, and the parcellation gets coarser.

### S.3.5 The (modified) Jaccard score

To quantitatively compare clusterings, we consider a variation of the standard Jaccard score,

$$\text{Sim}(\mathcal{C}_1, \mathcal{C}_2) = \frac{1}{\max(k, \ell)} \cdot \sum_{(i,j) \in \mathcal{E}} W_{ij}, \tag{S.3}$$

where $\mathcal{C}_1 = \{A_1, \ldots, A_k\}$ and $\mathcal{C}_2 = \{B_1, \ldots, B_\ell\}$ are two clusterings; above, $\mathcal{E} \subseteq \{1, \ldots, k\} \times \{1, \ldots, \ell\}$ is a maximum weighted edge covering in a (weighted) bipartite graph, where the vertices on a side of the graph correspond to the clusters in a clustering, and the edge weights $W_{ij}$, $i = 1, \ldots, m$, $j = 1, \ldots, \ell$, are the usual Jaccard scores given by $\frac{|A_i \cap B_j|}{|A_i \cup B_j|}$. The use of the edge covering here resolves various complications that arise when comparing clusterings of different sizes; the $\frac{1}{\max(k,\ell)}$ term in (S.3) can be thought of as a normalizing constant Finally, to compute the edge covering, we use the algorithm of [6].

### S.3.6 Expanded set of results

Here, we provide an expanded set of figures and tables from the experiments with (i) various penalty values of HP-CONCORD (*i.e.*, $\lambda_1$ and $\lambda_2$), (ii) two parts of the brain (left and right hemisphere), (iii) two clustering methods, taking in the partial correlation graph from HP-CONCORD as inputs (persistent homology and Louvain methods), and (iv) two clustering coarseness levels (more clusters and fewer clusters). For convenience, we outline where to find each of our results in the table below.

| Method | | Left hemisphere | | Right hemisphere | |
|---|---|---|---|---|---|
| | | Clusterings | Jaccard scores | Clusterings | Jaccard scores |
| **Persistent homology** | with $\varepsilon = 3$ (fewer clusters) | See Table 1 | See Table 9 | See Table 2 | See Table 10 |
| | with $\varepsilon = 0$ (more clusters) | See Table 3 | See Table 11 | See Table 4 | See Table 12 |
| **Louvain** | with $k = 0$ (fewer clusters) | See Table 5 | See Table 13 | See Table 6 | See Table 14 |
| | with $k = $ max. # clusters from Louvain | See Table 7 | See Table 15 | See Table 8 | See Table 16 |

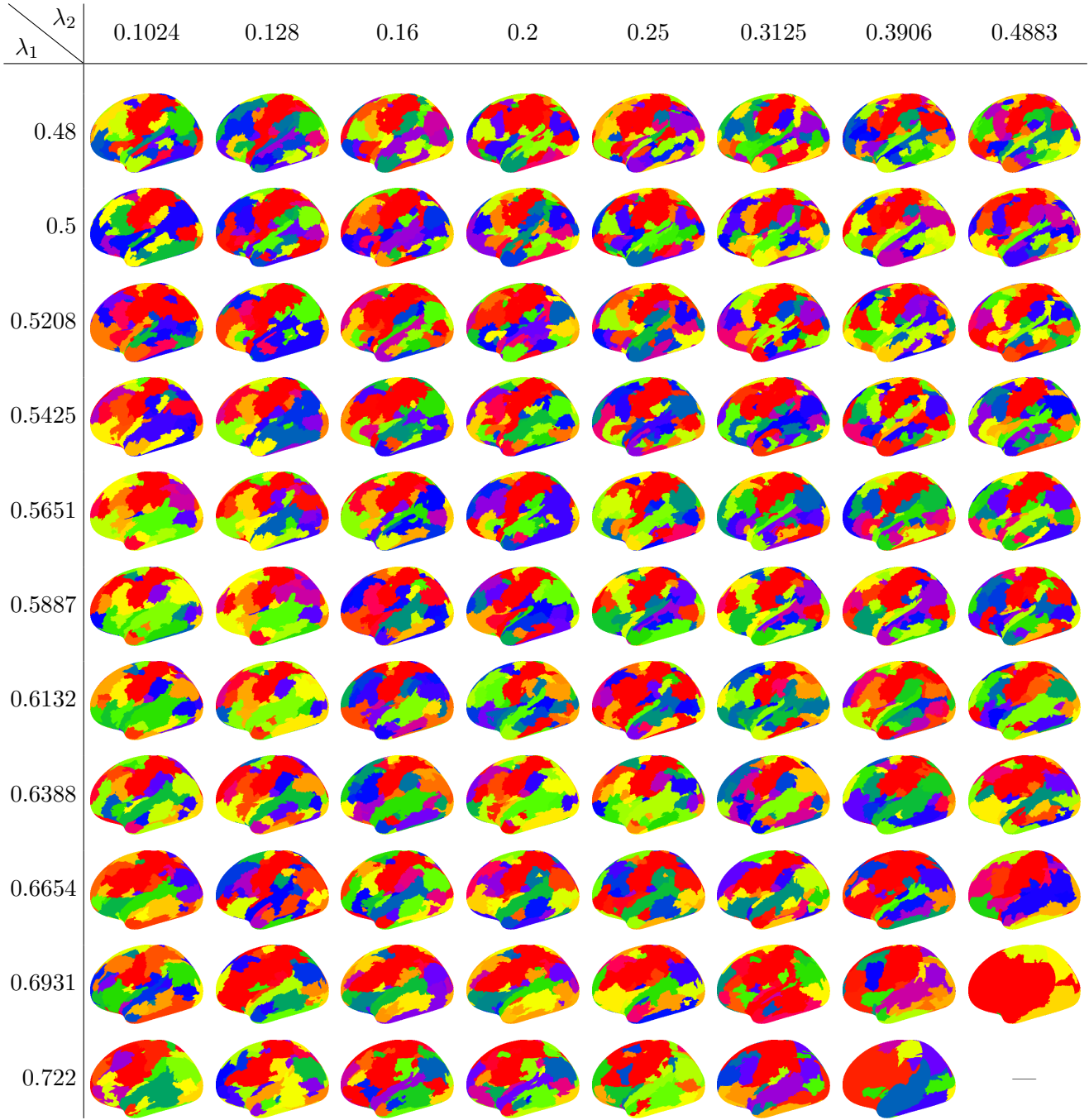

Table 1: The clusterings, for the *left* hemisphere, generated by HP-CONCORD followed by the *persistent homology* method, at the tuning parameter values: $\varepsilon = 3$ (generally corresponding to *fewer* clusters) as well as all the $\lambda_1, \lambda_2$ values we describe in Section 5. Table 9 presents the Jaccard scores (S.3) for these clusterings. "—", if present, indicates a degenerate clustering that puts either all the voxels into a single cluster or each voxel into its own cluster.

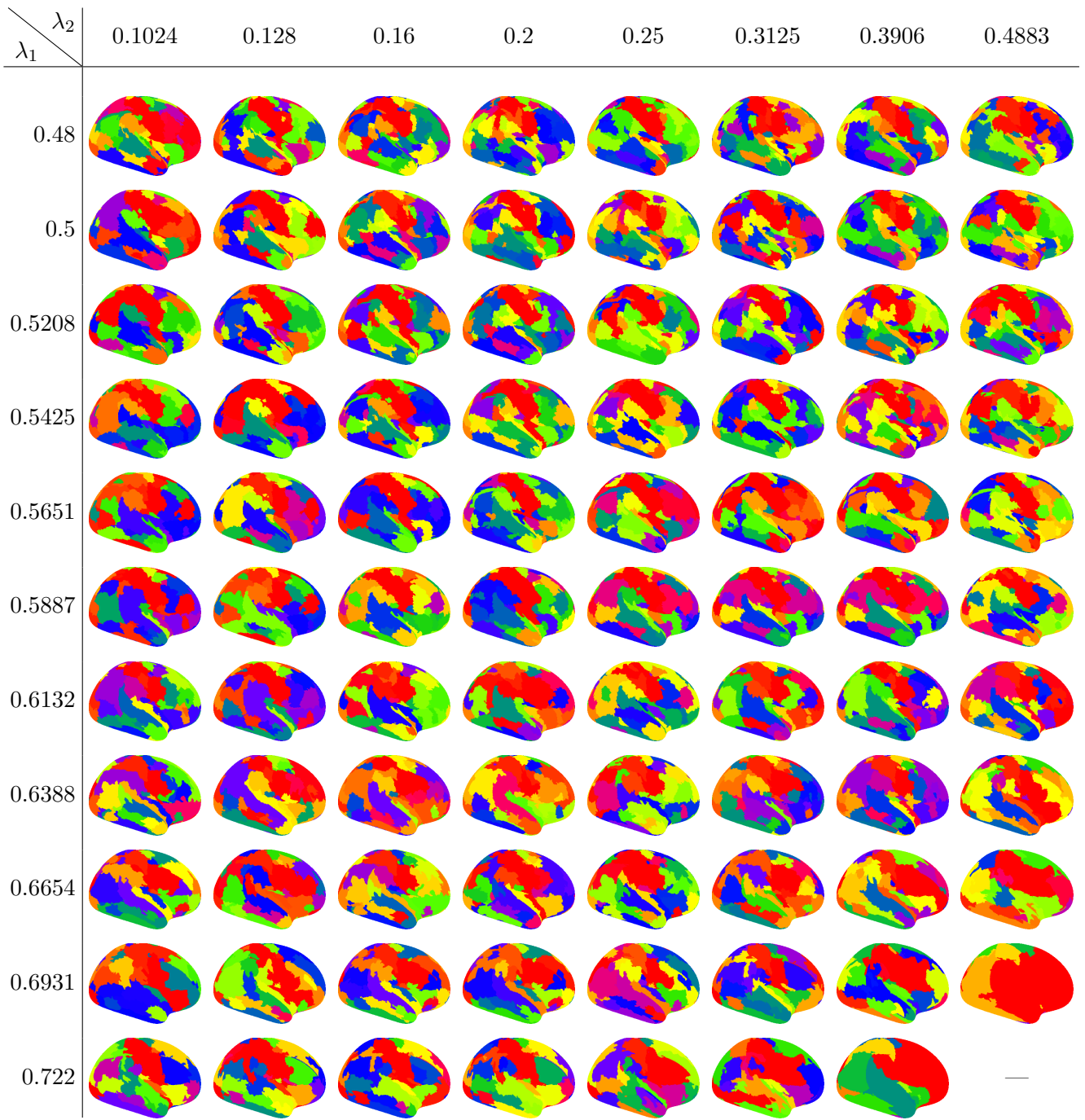

Table 2: The clusterings, for the *right* hemisphere, generated by HP-CONCORD followed by the *persistent homology* method, at the tuning parameter values: $\varepsilon = 3$ (generally corresponding to *fewer* clusters) as well as all the $\lambda_1, \lambda_2$ values we describe in Section 5. Table 10 presents the Jaccard scores (S.3) for these clusterings. "—", if present, indicates a degenerate clustering that puts either all the voxels into a single cluster or each voxel into its own cluster.

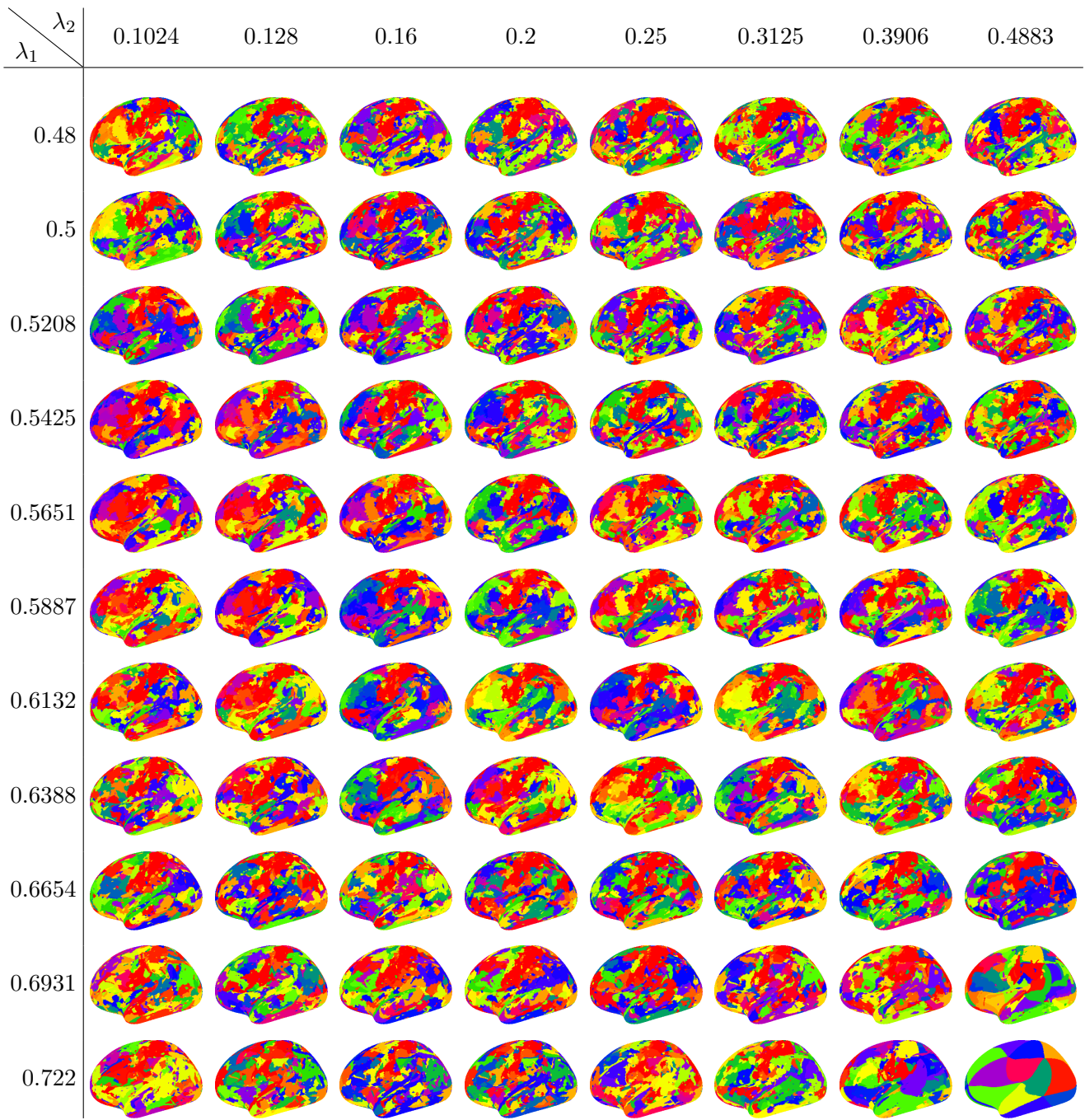

Table 3: The clusterings, for the *left* hemisphere, generated by HP-CONCORD followed by the *persistent homology* method, at the tuning parameter values: $\varepsilon = 0$ (generally corresponding to *more* clusters) as well as all the $\lambda_1, \lambda_2$ values we describe in Section 5. Table 11 presents the Jaccard scores (S.3) for these clusterings. "—", if present, indicates a degenerate clustering that puts either all the voxels into a single cluster or each voxel into its own cluster.

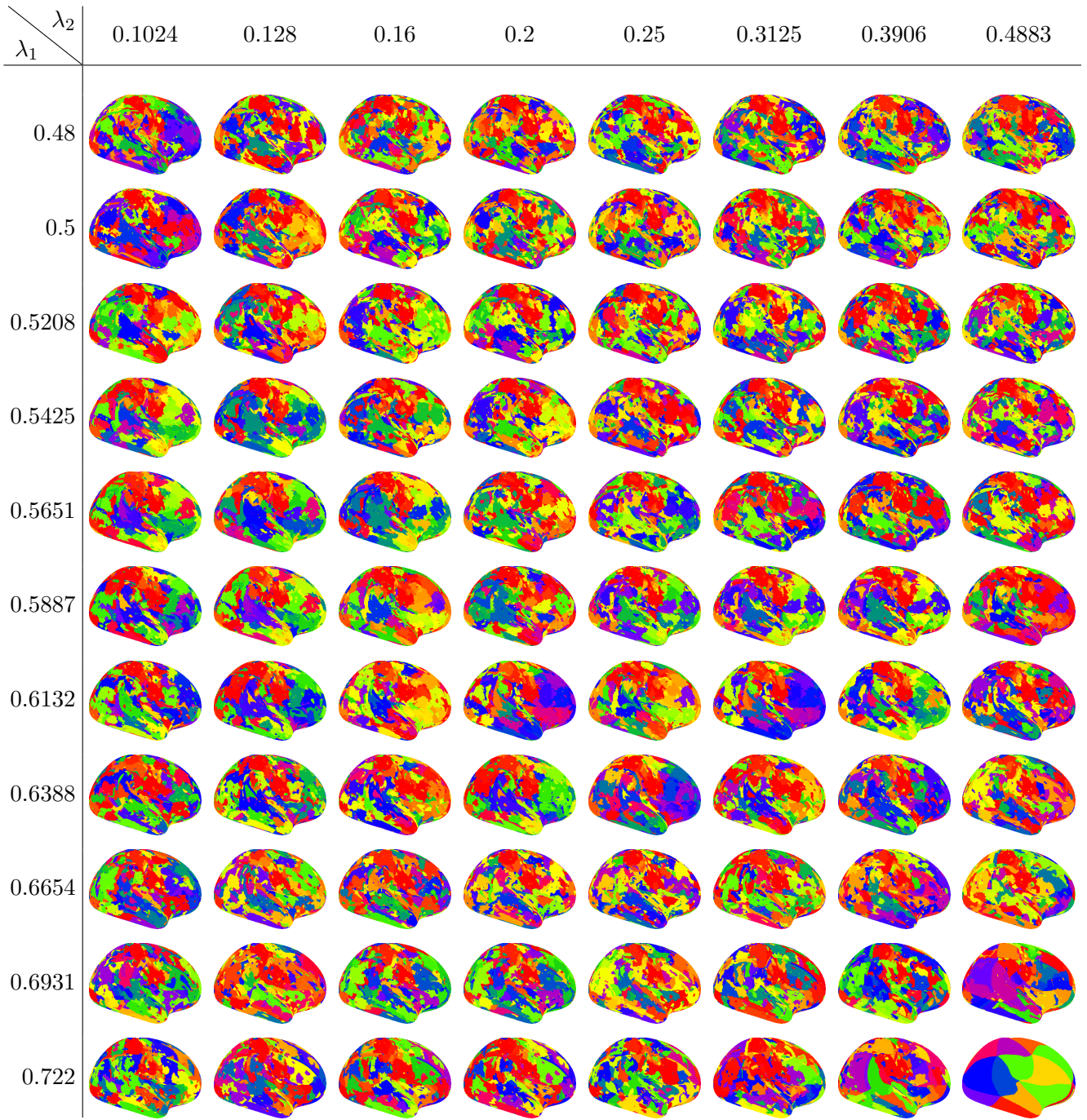

Table 4: The clusterings, for the *right* hemisphere, generated by HP-CONCORD followed by the *persistent homology* method, at the tuning parameter values: $\varepsilon = 0$ (generally corresponding to *more* clusters) as well as all the $\lambda_1, \lambda_2$ values we describe in Section 5. Table 12 presents the Jaccard scores (S.3) for these clusterings. "—", if present, indicates a degenerate clustering that puts either all the voxels into a single cluster or each voxel into its own cluster.

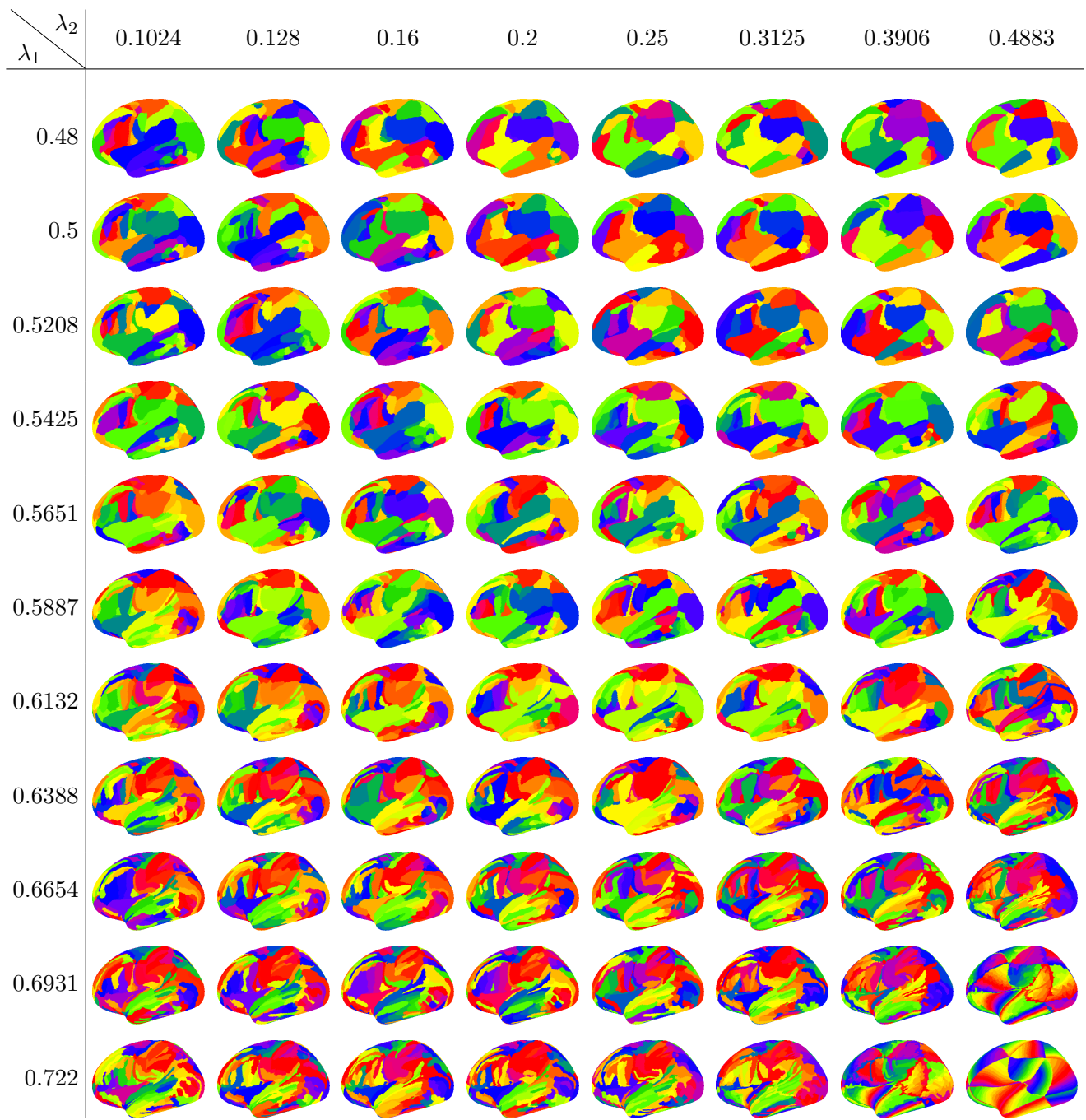

Table 5: The clusterings, for the *left* hemisphere, generated by HP-CONCORD followed by the *Louvain* method, at the tuning parameter values: $k = 0$ (generally corresponding to *fewer* clusters) as well as all the $\lambda_1, \lambda_2$ values we describe in Section 5. Table 13 presents the Jaccard scores (S.3) for these clusterings. "—", if present, indicates a degenerate clustering that puts either all the voxels into a single cluster or each voxel into its own cluster.

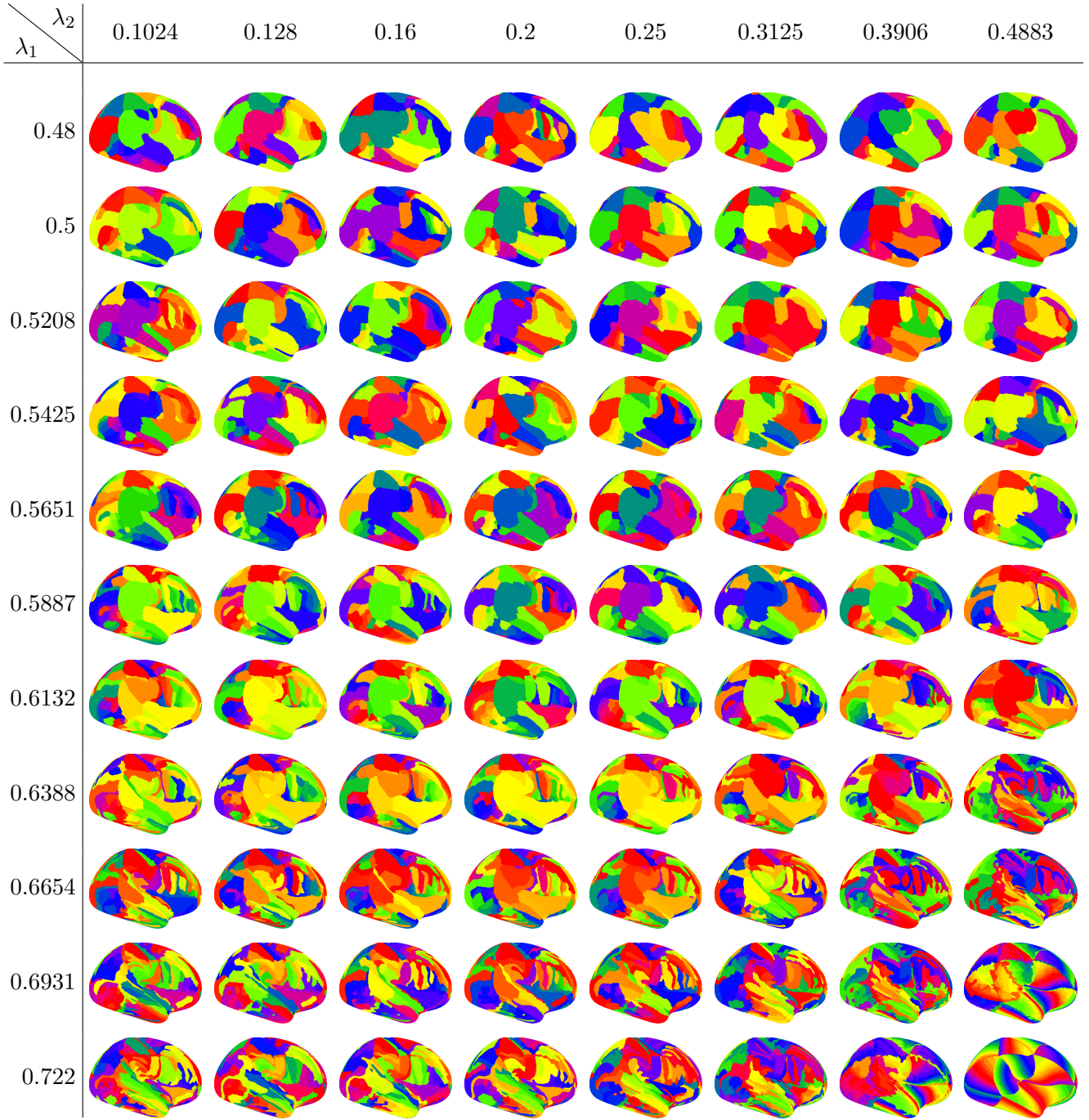

Table 6: The clusterings, for the *right* hemisphere, generated by HP-CONCORD followed by the *Louvain* method, at the tuning parameter values: $k = 0$ (generally corresponding to *fewer* clusters) as well as all the $\lambda_1, \lambda_2$ values we describe in Section 5. Table 14 presents the Jaccard scores (S.3) for these clusterings. "—", if present, indicates a degenerate clustering that puts either all the voxels into a single cluster or each voxel into its own cluster.

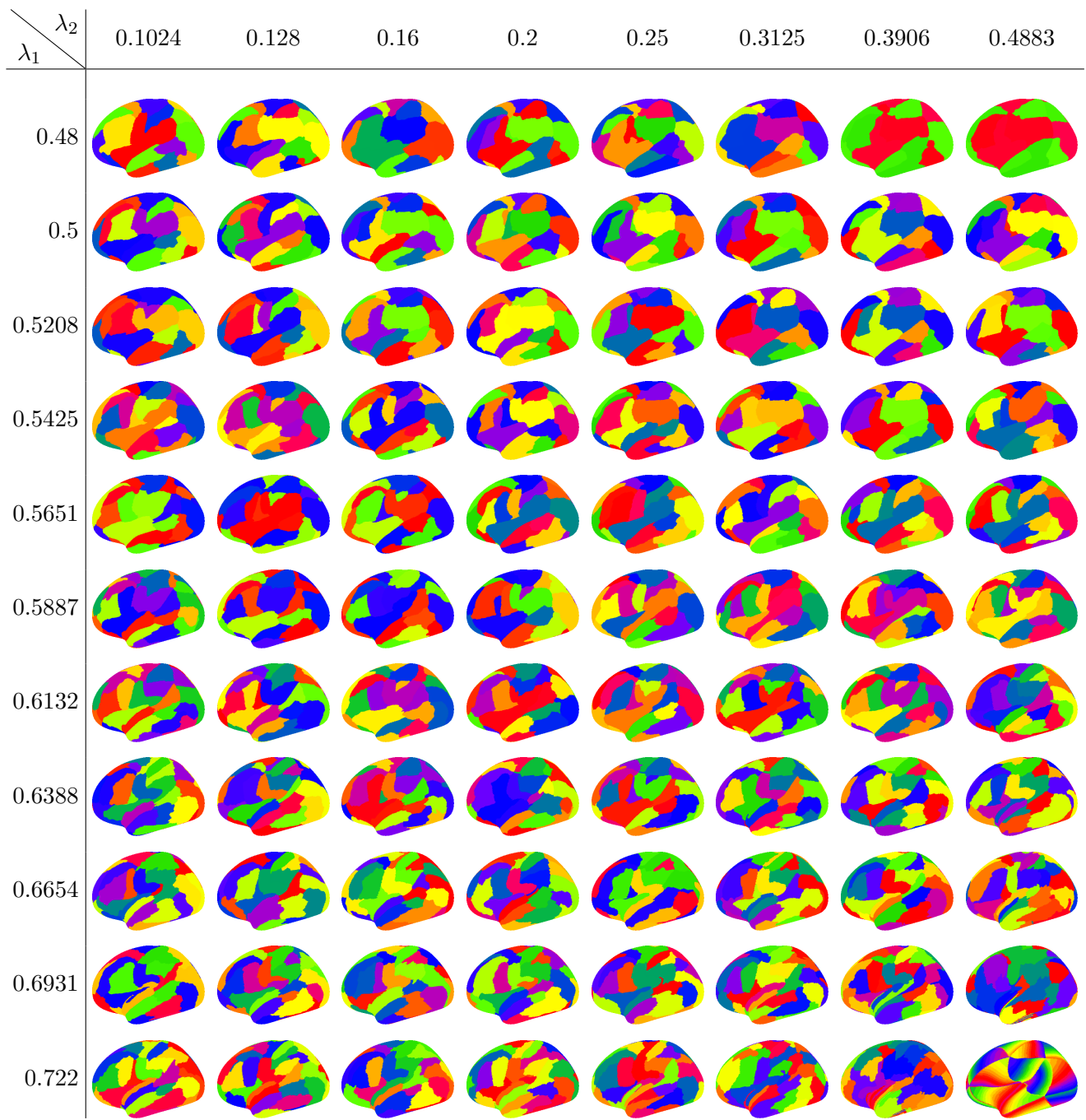

Table 7: The clusterings, for the *left* hemisphere, generated by HP-CONCORD followed by the *Louvain* method, at the tuning parameter values: the largest value of $k$ considered by Louvain (generally corresponding to *more* clusters) as well as all the $\lambda_1, \lambda_2$ values we describe in Section 5. Table 15 presents the Jaccard scores (S.3) for these clusterings. "—", if present, indicates a degenerate clustering that puts either all the voxels into a single cluster or each voxel into its own cluster.

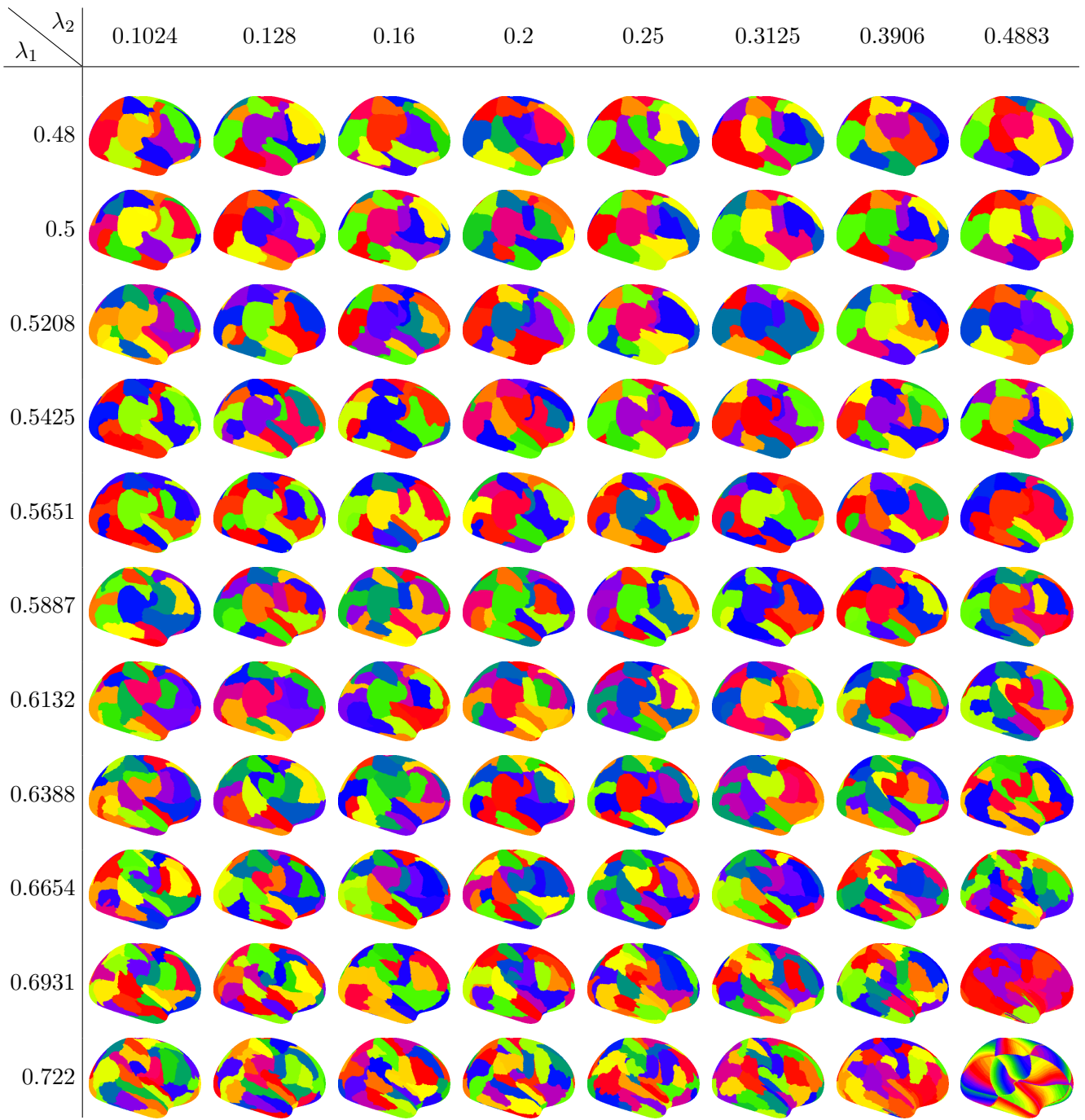

Table 8: The clusterings, for the *right* hemisphere, generated by HP-CONCORD followed by the *Louvain* method, at the tuning parameter values: the largest value of $k$ considered by Louvain (generally corresponding to *more* clusters) as well as all the $\lambda_1, \lambda_2$ values we describe in Section 5. Table 16 presents the Jaccard scores (S.3) for these clusterings. "—", if present, indicates a degenerate clustering that puts either all the voxels into a single cluster or each voxel into its own cluster.

| $\lambda_1$ \ $\lambda_2$ | 0.1024 | 0.128 | 0.16 | 0.2 | 0.25 | 0.3125 | 0.3906 | 0.4883 |
|---:|---|---|---|---|---|---|---|---|
| 0.48 | 0.2043 | 0.2199 | 0.2242 | 0.2326 | 0.2277 | 0.2422 | 0.2447 | 0.23 |
| 0.5 | 0.2112 | 0.224 | 0.2315 | 0.2329 | 0.2343 | 0.2283 | 0.2197 | 0.2185 |
| 0.5208 | 0.1964 | 0.1895 | 0.2264 | 0.2385 | 0.2317 | 0.2282 | 0.2348 | 0.2358 |
| 0.5425 | 0.1905 | 0.1972 | 0.1951 | 0.2181 | 0.2268 | 0.2295 | 0.2289 | 0.2255 |
| 0.5651 | 0.1833 | 0.197 | 0.1973 | 0.1981 | 0.2125 | 0.2268 | 0.2242 | 0.2213 |
| 0.5887 | 0.1838 | 0.1845 | 0.1992 | 0.2067 | 0.1953 | 0.2057 | 0.2078 | 0.2155 |
| 0.6132 | 0.1702 | 0.1752 | 0.198 | 0.1995 | 0.2121 | 0.2014 | 0.2036 | 0.1891 |
| 0.6388 | 0.1698 | 0.1693 | 0.1864 | 0.1837 | 0.1859 | 0.191 | 0.1831 | 0.1785 |
| 0.6654 | 0.1538 | 0.1854 | 0.1759 | 0.1701 | 0.1748 | 0.1844 | 0.1805 | 0.1467 |
| 0.6931 | 0.1652 | 0.1689 | 0.1664 | 0.1686 | 0.1722 | 0.162 | 0.1472 | 0.0516 |
| 0.722 | 0.1382 | 0.1536 | 0.1556 | 0.1536 | 0.1442 | 0.1394 | 0.0758 | — |

Table 9: The Jaccard scores (S.3) for the clusterings of the *left* hemisphere in Table 1, generated by HP-CONCORD followed by the *persistent homology* method, at the tuning parameter values: $\varepsilon = 3$ (generally corresponding to *fewer* clusters) as well as all the $\lambda_1, \lambda_2$ values we describe in Section 5. "—", if present, indicates a degenerate clustering that puts either all the voxels into a single cluster or each voxel into its own cluster.

| $\lambda_1$ \ $\lambda_2$ | 0.1024 | 0.128 | 0.16 | 0.2 | 0.25 | 0.3125 | 0.3906 | 0.4883 |
|---:|---|---|---|---|---|---|---|---|
| 0.48 | 0.2258 | 0.2315 | 0.2461 | 0.2279 | 0.2451 | 0.2436 | 0.2311 | 0.2431 |
| 0.5 | 0.2036 | 0.2245 | 0.2328 | 0.2326 | 0.2427 | 0.2314 | 0.2654 | 0.2528 |
| 0.5208 | 0.2255 | 0.2166 | 0.2317 | 0.2311 | 0.2427 | 0.2399 | 0.2381 | 0.2417 |
| 0.5425 | 0.21 | 0.2172 | 0.232 | 0.2355 | 0.2279 | 0.2299 | 0.245 | 0.2349 |
| 0.5651 | 0.2233 | 0.2182 | 0.2236 | 0.2341 | 0.2367 | 0.231 | 0.2286 | 0.2413 |
| 0.5887 | 0.2055 | 0.2187 | 0.2179 | 0.2369 | 0.2261 | 0.2321 | 0.2279 | 0.2067 |
| 0.6132 | 0.1843 | 0.2002 | 0.2245 | 0.2224 | 0.2113 | 0.219 | 0.2256 | 0.21 |
| 0.6388 | 0.1817 | 0.1843 | 0.2024 | 0.204 | 0.2154 | 0.2161 | 0.1981 | 0.1826 |
| 0.6654 | 0.1786 | 0.1678 | 0.1824 | 0.1891 | 0.1952 | 0.1749 | 0.1851 | 0.1273 |
| 0.6931 | 0.1652 | 0.1714 | 0.1686 | 0.1736 | 0.1714 | 0.1702 | 0.1284 | 0.061 |
| 0.722 | 0.1372 | 0.1562 | 0.162 | 0.1563 | 0.1364 | 0.1264 | 0.0875 | — |

Table 10: The Jaccard scores (S.3) for the clusterings of the *right* hemisphere, generated by HP-CONCORD followed by the *persistent homology* method, at the tuning parameter values: $\varepsilon = 3$ (generally corresponding to *fewer* clusters) as well as all the $\lambda_1, \lambda_2$ values we describe in Section 5. "—", if present, indicates a degenerate clustering that puts either all the voxels into a single cluster or each voxel into its own cluster.

| $\lambda_1$ \ $\lambda_2$ | 0.1024 | 0.128 | 0.16 | 0.2 | 0.25 | 0.3125 | 0.3906 | 0.4883 |
|---|---|---|---|---|---|---|---|---|
| 0.48   | 0.0507 | 0.051  | 0.0527 | 0.0532 | 0.0511 | 0.0503 | 0.051  | 0.0518 |
| 0.5    | 0.053  | 0.0518 | 0.052  | 0.0519 | 0.0526 | 0.0531 | 0.0517 | 0.0516 |
| 0.5208 | 0.0517 | 0.0519 | 0.0527 | 0.0517 | 0.0524 | 0.0526 | 0.0522 | 0.0536 |
| 0.5425 | 0.0522 | 0.0519 | 0.0509 | 0.0516 | 0.0516 | 0.0514 | 0.0532 | 0.0533 |
| 0.5651 | 0.0514 | 0.0524 | 0.0512 | 0.0528 | 0.0529 | 0.0518 | 0.0518 | 0.0533 |
| 0.5887 | 0.0498 | 0.0524 | 0.0534 | 0.0521 | 0.0522 | 0.0521 | 0.0526 | 0.0532 |
| 0.6132 | 0.0504 | 0.0501 | 0.0531 | 0.052  | 0.0523 | 0.0529 | 0.0523 | 0.0505 |
| 0.6388 | 0.053  | 0.052  | 0.0494 | 0.0502 | 0.0517 | 0.0502 | 0.0516 | 0.0543 |
| 0.6654 | 0.0526 | 0.0529 | 0.0536 | 0.0533 | 0.0537 | 0.0506 | 0.0535 | 0.0558 |
| 0.6931 | 0.0529 | 0.054  | 0.0518 | 0.052  | 0.0532 | 0.0543 | 0.0566 | 0.0815 |
| 0.722  | 0.0549 | 0.0528 | 0.0525 | 0.0534 | 0.056  | 0.0561 | 0.0718 | 0.0884 |

Table 11: The Jaccard scores (S.3) for the clusterings of the *left* hemisphere, generated by HP-CONCORD followed by the *persistent homology* method, at the tuning parameter values: $\varepsilon = 0$ (generally corresponding to *more* clusters) as well as all the $\lambda_1, \lambda_2$ values we describe in Section 5. "—", if present, indicates a degenerate clustering that puts either all the voxels into a single cluster or each voxel into its own cluster.

| $\lambda_1$ \ $\lambda_2$ | 0.1024 | 0.128 | 0.16 | 0.2 | 0.25 | 0.3125 | 0.3906 | 0.4883 |
|---|---|---|---|---|---|---|---|---|
| 0.48   | 0.0532 | 0.0506 | 0.0519 | 0.0516 | 0.0522 | 0.0521 | 0.0516 | 0.0508 |
| 0.5    | 0.0538 | 0.0536 | 0.0513 | 0.0512 | 0.0515 | 0.0525 | 0.0524 | 0.0506 |
| 0.5208 | 0.052  | 0.0519 | 0.0521 | 0.0509 | 0.0527 | 0.0517 | 0.0522 | 0.0509 |
| 0.5425 | 0.0505 | 0.0523 | 0.0528 | 0.0532 | 0.0511 | 0.0529 | 0.0526 | 0.0522 |
| 0.5651 | 0.0523 | 0.0501 | 0.0513 | 0.0513 | 0.053  | 0.0521 | 0.0512 | 0.0528 |
| 0.5887 | 0.0516 | 0.0528 | 0.0504 | 0.0515 | 0.0518 | 0.0515 | 0.0523 | 0.0511 |
| 0.6132 | 0.0505 | 0.0517 | 0.0525 | 0.0534 | 0.0511 | 0.0516 | 0.0543 | 0.0534 |
| 0.6388 | 0.0514 | 0.0543 | 0.0516 | 0.0522 | 0.0519 | 0.0533 | 0.0532 | 0.0544 |
| 0.6654 | 0.0544 | 0.0528 | 0.0514 | 0.0518 | 0.0525 | 0.0529 | 0.0565 | 0.061  |
| 0.6931 | 0.0527 | 0.0555 | 0.0525 | 0.0528 | 0.055  | 0.0529 | 0.0597 | 0.0845 |
| 0.722  | 0.0535 | 0.0524 | 0.0535 | 0.0527 | 0.0553 | 0.0566 | 0.0698 | 0.0918 |

Table 12: The Jaccard scores (S.3) for the clusterings of the *right* hemisphere, generated by HP-CONCORD followed by the *persistent homology* method, at the tuning parameter values: $\varepsilon = 0$ (generally corresponding to *more* clusters) as well as all the $\lambda_1, \lambda_2$ values we describe in Section 5. "—", if present, indicates a degenerate clustering that puts either all the voxels into a single cluster or each voxel into its own cluster.

| $\lambda_1$ \ $\lambda_2$ | 0.1024 | 0.128 | 0.16 | 0.2 | 0.25 | 0.3125 | 0.3906 | 0.4883 |
|---|---|---|---|---|---|---|---|---|
| 0.48 | 0.1069 | 0.1101 | 0.0956 | 0.1042 | 0.1015 | 0.0958 | 0.0901 | 0.0905 |
| 0.5 | 0.1123 | 0.1076 | 0.1065 | 0.102 | 0.1089 | 0.1053 | 0.1007 | 0.107 |
| 0.5208 | 0.1097 | 0.111 | 0.1092 | 0.1096 | 0.1065 | 0.0982 | 0.1001 | 0.105 |
| 0.5425 | 0.1313 | 0.1123 | 0.1148 | 0.1085 | 0.1166 | 0.1143 | 0.1065 | 0.117 |
| 0.5651 | 0.1258 | 0.1216 | 0.1134 | 0.1167 | 0.1164 | 0.1097 | 0.1151 | 0.12 |
| 0.5887 | 0.129 | 0.1228 | 0.1233 | 0.1091 | 0.1203 | 0.1205 | 0.1238 | 0.1188 |
| 0.6132 | 0.1337 | 0.1294 | 0.1298 | 0.1289 | 0.1185 | 0.1285 | 0.1231 | 0.1455 |
| 0.6388 | 0.1477 | 0.1368 | 0.1363 | 0.1296 | 0.131 | 0.1344 | 0.1473 | 0.1517 |
| 0.6654 | 0.1486 | 0.1486 | 0.1458 | 0.1405 | 0.1488 | 0.1534 | 0.1486 | 0.1583 |
| 0.6931 | 0.1469 | 0.1453 | 0.1512 | 0.1483 | 0.146 | 0.1627 | 0.1706 | 0.0273 |
| 0.722 | 0.1581 | 0.1608 | 0.1557 | 0.1608 | 0.1661 | 0.1779 | 0.0461 | 0.0061 |

Table 13: The Jaccard scores (S.3) for the clusterings of the *left* hemisphere, generated by HP-CONCORD followed by the *Louvain* method, at the tuning parameter values: $k = 0$ (generally corresponding to *fewer* clusters) as well as all the $\lambda_1, \lambda_2$ values we describe in Section 5. "—", if present, indicates a degenerate clustering that puts either all the voxels into a single cluster or each voxel into its own cluster.

| $\lambda_1$ \ $\lambda_2$ | 0.1024 | 0.128 | 0.16 | 0.2 | 0.25 | 0.3125 | 0.3906 | 0.4883 |
|---|---|---|---|---|---|---|---|---|
| 0.48 | 0.1105 | 0.1014 | 0.1034 | 0.1042 | 0.0988 | 0.0976 | 0.0976 | 0.0935 |
| 0.5 | 0.1084 | 0.1064 | 0.1011 | 0.1105 | 0.1003 | 0.1012 | 0.0992 | 0.1022 |
| 0.5208 | 0.1263 | 0.1059 | 0.1195 | 0.1056 | 0.0995 | 0.1059 | 0.1 | 0.1051 |
| 0.5425 | 0.122 | 0.1167 | 0.1113 | 0.1111 | 0.0997 | 0.1085 | 0.1125 | 0.0997 |
| 0.5651 | 0.1219 | 0.1212 | 0.1144 | 0.1022 | 0.1044 | 0.1109 | 0.1029 | 0.1189 |
| 0.5887 | 0.1218 | 0.1205 | 0.1184 | 0.1219 | 0.1159 | 0.1202 | 0.1183 | 0.135 |
| 0.6132 | 0.132 | 0.1259 | 0.1339 | 0.1265 | 0.1269 | 0.124 | 0.1294 | 0.1361 |
| 0.6388 | 0.1362 | 0.1364 | 0.1289 | 0.1286 | 0.1318 | 0.1279 | 0.1357 | 0.158 |
| 0.6654 | 0.1483 | 0.1451 | 0.1428 | 0.142 | 0.1438 | 0.1498 | 0.1626 | 0.1675 |
| 0.6931 | 0.1518 | 0.1552 | 0.1451 | 0.1473 | 0.1552 | 0.1671 | 0.1736 | 0.027 |
| 0.722 | 0.1648 | 0.1725 | 0.1556 | 0.1607 | 0.1643 | 0.1758 | 0.0482 | 0.0061 |

Table 14: The Jaccard scores (S.3) for the clusterings of the *right* hemisphere, generated by HP-CONCORD followed by the *Louvain* method, at the tuning parameter values: $k = 0$ (generally corresponding to *fewer* clusters) as well as all the $\lambda_1, \lambda_2$ values we describe in Section 5. "—", if present, indicates a degenerate clustering that puts either all the voxels into a single cluster or each voxel into its own cluster.

| $\lambda_1$ \ $\lambda_2$ | 0.1024 | 0.128 | 0.16 | 0.2 | 0.25 | 0.3125 | 0.3906 | 0.4883 |
|---|---|---|---|---|---|---|---|---|
| 0.48   | 0.1678 | 0.1666 | 0.154  | 0.14   | 0.1297 | 0.135  | 0.1311 | 0.1284 |
| 0.5    | 0.1632 | 0.1778 | 0.1595 | 0.1515 | 0.1537 | 0.1454 | 0.128  | 0.1422 |
| 0.5208 | 0.1578 | 0.1719 | 0.1589 | 0.1663 | 0.1604 | 0.1572 | 0.145  | 0.1609 |
| 0.5425 | 0.1538 | 0.1572 | 0.166  | 0.1542 | 0.1702 | 0.1689 | 0.1684 | 0.1569 |
| 0.5651 | 0.1503 | 0.1602 | 0.1541 | 0.1473 | 0.1561 | 0.1587 | 0.1502 | 0.155  |
| 0.5887 | 0.1526 | 0.158  | 0.1537 | 0.1622 | 0.1564 | 0.1547 | 0.149  | 0.1338 |
| 0.6132 | 0.1438 | 0.1425 | 0.154  | 0.1487 | 0.151  | 0.1489 | 0.1327 | 0.1191 |
| 0.6388 | 0.1414 | 0.1453 | 0.134  | 0.1431 | 0.1393 | 0.1357 | 0.1238 | 0.0967 |
| 0.6654 | 0.1252 | 0.1263 | 0.1403 | 0.1301 | 0.1279 | 0.1196 | 0.0987 | 0.0653 |
| 0.6931 | 0.1137 | 0.1159 | 0.1161 | 0.1163 | 0.109  | 0.0937 | 0.0701 | 0.0216 |
| 0.722  | 0.1008 | 0.1005 | 0.1015 | 0.0961 | 0.0891 | 0.0679 | 0.0298 | 0.0061 |

Table 15: The Jaccard scores (S.3) for the clusterings of the *left* hemisphere, generated by HP-CONCORD followed by the *Louvain* method, at the tuning parameter values: the largest value of $k$ considered by Louvain (generally corresponding to *more* clusters) as well as all the $\lambda_1, \lambda_2$ values we describe in Section 5. "—", if present, indicates a degenerate clustering that puts either all the voxels into a single cluster or each voxel into its own cluster.

| $\lambda_1$ \ $\lambda_2$ | 0.1024 | 0.128 | 0.16 | 0.2 | 0.25 | 0.3125 | 0.3906 | 0.4883 |
|---|---|---|---|---|---|---|---|---|
| 0.48   | 0.1719 | 0.1697 | 0.1633 | 0.1763 | 0.1499 | 0.1475 | 0.1442 | 0.1365 |
| 0.5    | 0.1689 | 0.1675 | 0.167  | 0.17   | 0.1661 | 0.1587 | 0.1411 | 0.1729 |
| 0.5208 | 0.1651 | 0.1581 | 0.1808 | 0.1694 | 0.1655 | 0.1528 | 0.153  | 0.1512 |
| 0.5425 | 0.1697 | 0.1556 | 0.1634 | 0.1637 | 0.1591 | 0.1581 | 0.1857 | 0.1598 |
| 0.5651 | 0.1651 | 0.1663 | 0.1509 | 0.1554 | 0.1567 | 0.1542 | 0.151  | 0.1416 |
| 0.5887 | 0.1492 | 0.1602 | 0.1635 | 0.1541 | 0.1512 | 0.1586 | 0.1506 | 0.1536 |
| 0.6132 | 0.1474 | 0.1596 | 0.1586 | 0.1593 | 0.1649 | 0.1548 | 0.1436 | 0.1168 |
| 0.6388 | 0.1321 | 0.1337 | 0.1495 | 0.1502 | 0.1458 | 0.1272 | 0.1188 | 0.0938 |
| 0.6654 | 0.119  | 0.1203 | 0.1233 | 0.1221 | 0.1185 | 0.1125 | 0.0973 | 0.0635 |
| 0.6931 | 0.112  | 0.1136 | 0.1128 | 0.111  | 0.107  | 0.0932 | 0.0694 | 0.0213 |
| 0.722  | 0.0943 | 0.098  | 0.0994 | 0.0937 | 0.0832 | 0.0672 | 0.0299 | 0.0061 |

Table 16: The Jaccard scores (S.3) for the clusterings of the *right* hemisphere, generated by HP-CONCORD followed by the *Louvain* method, at the tuning parameter values: the largest value of $k$ considered by Louvain (generally corresponding to *more* clusters) as well as all the $\lambda_1, \lambda_2$ values we describe in Section 5. "—", if present, indicates a degenerate clustering that puts either all the voxels into a single cluster or each voxel into its own cluster.